\documentclass{article}

\usepackage{microtype}
\usepackage{graphicx}
\usepackage{subcaption}
\usepackage{booktabs} 

\usepackage{hyperref}

\usepackage[accepted]{icml2026}

\usepackage{amsmath}
\usepackage{amssymb}
\usepackage{mathtools}
\usepackage{amsthm}

\usepackage[capitalize,noabbrev]{cleveref}

\theoremstyle{plain}

\theoremstyle{definition}

\theoremstyle{remark}

\usepackage{url}
\usepackage{multirow}
\usepackage{tabularx}
\usepackage[table]{xcolor}
\usepackage{tcolorbox}
\usepackage{makecell}
\usepackage{amsfonts}
\usepackage{nicefrac}
\usepackage{microtype}
\usepackage{times}
\usepackage{xspace}
\usepackage{amsmath}
\usepackage{enumitem}
\usepackage{caption}
\usepackage{listings}
\usepackage{verbatim}
\usepackage{titletoc}

\definecolor{highlightrow_yellow}{HTML}{FFFFC7} 
\definecolor{highlightrow}{HTML}{E6F3FF}

\icmltitlerunning{Optimizing Inference-Time Compute for Medical Reasoning via Uncertainty Quantification}

\begin{document}

\twocolumn[
  \icmltitle{Optimizing Inference-Time Compute for Medical Reasoning via Uncertainty Quantification}



  \icmlsetsymbol{equal}{*}

  \begin{icmlauthorlist}
    \icmlauthor{Shaohao Rui}{sjtu,sii}
    \icmlauthor{Kaitao Chen}{fudan,sail}
    \icmlauthor{Weijie Ma}{sii,fudan}
    \icmlauthor{Xiaosong Wang}{sii,sail}
  \end{icmlauthorlist}

  \icmlaffiliation{sjtu}{Shanghai Jiao Tong University, Shanghai, China}
  \icmlaffiliation{sii}{Shanghai Innovation Institute, Shanghai, China}
  \icmlaffiliation{fudan}{Fudan University, Shanghai, China}
  \icmlaffiliation{sail}{Shanghai AI Laboratory, Shanghai, China}

  \icmlcorrespondingauthor{Xiaosong Wang}{wangxiaosong@pjlab.org.cn}

  \icmlkeywords{Machine Learning, ICML}

  \vskip 0.3in
]



\printAffiliationsAndNotice{}  

\begin{abstract}
Extended Chain-of-Thought (CoT) reasoning has significantly bolstered the capabilities of medical large language models (LLMs). However, current models exhibit static computational expenditure, applying lengthy reasoning processes indiscriminately to both simple queries and complex diagnostic cases. This inefficiency is particularly prohibitive in real-world healthcare, where clinical scenarios range from time-sensitive emergencies requiring rapid response to intricate pathologies demanding deep analysis. To address this, we propose \textbf{AdaThink-Med}\footnote{Code available at \url{https://github.com/shaohao011/AdaThinkMed}}, an end-to-end framework for adaptive reasoning via uncertainty-guided length calibration. Although the underlying mechanism is generalizable, we demonstrate its critical value in the medical domain, where balancing inference latency with diagnostic precision is paramount. AdaThink-Med leverages entropy-based uncertainty estimation within reinforcement fine-tuning to dynamically shape reward signals: it penalizes verbosity for high-confidence correct answers (e.g., straightforward knowledge retrieval) while incentivizing extended exploration for uncertain or ambiguous scenarios. Across six medical benchmarks, AdaThink-Med reduces inference token consumption by \textbf{4.7$\times$ to 6.4$\times$} on Qwen and Llama architectures, respectively, with minimal performance trade-offs. Notably, our reward design naturally produces distinct ``non-thinking'' and ``thinking'' modes within a single model, enabling efficient allocation of computational resources without any external router or classifier.
\end{abstract}

\section{Introduction}
Recent advances in Chain-of-Thought (CoT) reasoning~\citep{wei2022chain} and reinforcement learning~\citep{suzgun2022challenging,zhou2022least,chen2024language,guo2025deepseek} have significantly improved the reasoning abilities of LLMs with leveraging extended trial-and-error processes and reflective thinking---also known as inference time scaling. Such progress has inspired new paradigms for building general-purpose medical AI systems. Through high-quality long CoT data for supervised fine-tuning or rule-based reinforcement learning, substantial gains have been made in complex clinical tasks including differential diagnosis, surgical planning, and prognosis prediction~\citep{chen2024huatuogpt-o1,huang2025m1,liu2025beyond,rui2026cardiocot,rui2025improving,chen2025think}. 

However, such improvements often come at the cost of significantly increased lengthy outputs, thereby raising inference time and costs. For example, when answering a simple question like ``1 + 1 = ?'', models such as DeepSeek-R1~\citep{guo2025deepseek} have been observed to produce outputs as long as 170 tokens. In parallel, another line of research has identified the \textit{overthinking} phenomenon~\citep{chen2024not,luo2025o1}, where reasoning models generate unnecessarily long chains of thought, which can even degrade performance on simple questions~\citep{su2025between}. Moreover,~\citep{zhang2025synapseroute} reports that up to 58\% of questions in existing medical benchmarks can be accurately answered with short, concise outputs---without resorting to expensive extended reasoning. These findings collectively underscore the urgent need to improve the efficiency of reasoning models, especially if they are to be deployed effectively in real-world applications.

Most existing work aims to improve the efficiency of general reasoning models by applying length-based reward, which penalizes outputs with long CoT. Specifically, \citep{team2025kimi,yu2025dapo} encourages shorter outputs across the board, while another line of work \citep{shen2025dast,yi2025shorterbetter,yeo2025demystifying} penalizes long outputs when the answer is correct but encourages longer outputs for incorrect ones to allow more exploration. However, these methods rely solely on correctness and largely ignore the underlying {problem difficulty}, which limits their ability to achieve optimal efficiency improvements. Ideally, from a human perspective, computational resources should be allocated adaptively, i.e., spending more effort on complex questions and less on simpler ones. Recent studies further suggest that difficult questions often benefit from longer CoT to maintain high performance, whereas simple problems can be solved with much shorter outputs or even without explicit reasoning~\citep{liu2025learn,ma2025reasoning,jiang2025think,fang2025thinkless}. In the medical domain, pioneering work \citep{zhang2025synapseroute} decomposes the reasoning process into thinking and non-thinking modes based on the estimated difficulty of each question. However, it relies on manual annotations to train a difficulty classifier, introducing scalability bottlenecks and potential labeling bias.

Furthermore, recent studies highlight a persistent trade-off between performance and inference cost in hybrid adaptive models. For instance, Qwen3 exhibits notable performance degradation in its hybrid architecture particularly for non-thinking tasks—compared to specialized dual-model setups, prompting a strategic reversion to separate models for reasoning-intensive versus routine queries~\citep{qwen3technicalreport,alibaba2025qwen3}. Similarly, while DeepSeekV3.1~\citep{deepseekai2024deepseekv3technicalreport} and GPT-5 adopt hybrid architectures, they necessitate vast additional data to mitigate performance loss, yet user feedback still favors the specialized experience of dual models. Consequently, achieving an optimal balance within a single unified model remains challenging, underscoring the urgent need for more effective optimization strategies.

Motivated by the above observations, we propose AdaThink-Med, the first end-to-end framework designed to improve adaptive thinking capabilities in medical language models by explicitly incorporating problem difficulty. The central idea of AdaThink-Med is to enable the model to generate outputs of varying lengths depending on the difficulty of the input question, producing concise answers for simple cases and long reasoning chains for complex ones. During training, for each question, we generate multiple outputs and evaluate both their accuracy and uncertainty to derive a comprehensive measure of difficulty. This difficulty-aware signal is then incorporated into the reinforcement learning process, guiding the model to generate shorter outputs for low-difficulty with high-confidence right answer. At inference time, AdaThink-Med automatically estimates the difficulty of incoming questions and generates outputs with an appropriate amount of reasoning.

We conduct comprehensive experiments across both rule-based and model-based RL paradigms, and conduct the evaluation on six public medical reasoning benchmarks spanning various difficulty levels. Results show that AdaThink-Med consistently achieves better efficiency compared to existing medical language models and length-based output control methods. 

The contributions of this work are fourfold:
\begin{itemize}[leftmargin=*]
    \item We propose \textbf{AdaThink-Med}, an end-to-end framework that allows a single medical LLM to switch between reasoning modes adaptively based on problem difficulty, without any external router or classifier.
    \item We introduce an entropy-guided length calibration mechanism whose reward design naturally produces distinct ``non-thinking'' (direct response) and ``thinking'' (extended reasoning) modes within one model.
    \item We demonstrate the framework's utility in dataset selection, facilitating the efficient extraction of high-quality training subsets from large-scale corpora.
    \item We achieve state-of-the-art performance across six medical benchmarks, delivering a superior trade-off between diagnostic accuracy and computational cost.
\end{itemize}

\section{Related Works}
\noindent\textbf{Efficient Reasoning via Reinforcement Learning.} Efficient reasoning seeks to mitigate the ``overthinking phenomenon'' by optimizing the trade-off between accuracy and computational cost~\citep{sui2025stop}. Prior works, such as Kimi1.5~\citep{team2025kimi}, DAPO~\citep{yu2025dapo}, and ShortBetter~\citep{yi2025shorterbetter}, primarily rely on greedy length penalties or minimal-length targeting. However, these strategies often lead to reward hacking and performance degradation, as they fail to incentivize extended reasoning for complex problems. While approaches like CosFn~\citep{yeo2025demystifying} and DAST~\citep{shen2025dast} attempt to address this by incorporating correctness or static difficulty, they generally overlook the inherent uncertainty in reasoning or rely on inflexible offline training. \textbf{In contrast}, our method introduces an entropy-based uncertainty measure for dynamic difficulty estimation within an online learning framework, enabling adaptive thinking that continuously evolves with the model's capabilities.

\noindent\textbf{Entropy-Guided Reasoning in LLMs.}
Recent research has increasingly leveraged entropy as a proxy for uncertainty to steer reasoning processes. Notable approaches include using semantic entropy for sample weighting in RL~\citep{chen2025seed}, dynamic inference termination~\citep{xu2025adaptive}, or reasoning trace compression~\citep{li2025compressing,zhu2025entropy}. Distinct from these works which primarily focus on re-weighting or explicit pruning, \textbf{AdaThink-Med} proposes a specific entropy-based difficulty estimator that fuses correctness with uncertainty. We integrate this estimator into a difficulty-aware length reward mechanism during reinforcement fine-tuning. This design dynamically penalizes redundancy for easy queries while incentivizing exploration for hard ones, rather than simply shortening outputs. To our knowledge, this represents the first application of such an entropy-guided mechanism to medical RL, where our reward design naturally produces reasoning mode separation and additionally enables effective dataset selection.

\begin{figure*}[t] 
    \centering
    \includegraphics[height=0.45\linewidth]{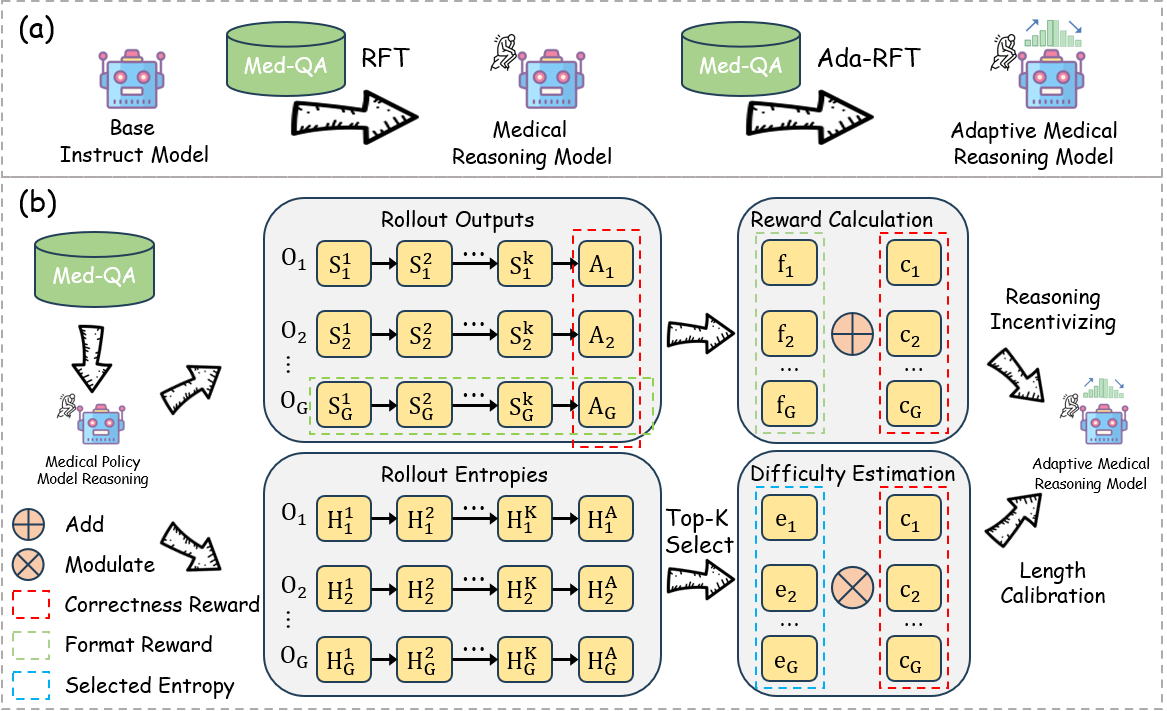}
    \caption{\textbf{Overview of the AdaThink-Med framework.} (a) \textbf{Staged Adaptive Training:} The training process consists of two stages. First, a Base Instruct Model is trained via standard Reinforcement Fine-Tuning (RFT) to establish medical reasoning capabilities. Subsequently, the model undergoes Adaptive RFT (Ada-RFT) to optimize reasoning efficiency. (b) \textbf{Uncertainty-Guided Difficulty Estimation and Length Calibration:} During the adaptive phase, the model generates multiple rollout outputs ($O_1 \dots O_G$). The framework simultaneously calculates standard rewards (correctness and format) and estimates output uncertainty via \textbf{Top-K token entropies}. By combining correctness ($c_i$) and uncertainty ($e_i$), the system estimates problem difficulty, which dynamically calibrates the length reward to incentivize concise answers for simple questions and extended reasoning for complex ones.}
    \label{fig:overview}
\end{figure*}

\section{Method}

\subsection{Overview of AdaThink-Med}
The core objective of AdaThink-Med is to transform a static medical reasoning model into an adaptive model that dynamically allocates computational resources based on clinical complexity. Formally, we aim to maximize the expected reward while minimizing the inference cost, regularized by a reference policy (see Appendix~\ref{app:preliminaries}).

Unlike prior approaches that impose uniform length penalties, AdaThink-Med introduces an uncertainty-guided length calibration mechanism. As illustrated in Figure~\ref{fig:overview}, the framework operates in two phases. First, for a given medical query, the model generates multiple reasoning trajectories. We estimate the problem difficulty by jointly analyzing the correctness of these responses and their intrinsic predictive uncertainty, quantified via entropy. Second, during Adaptive Reinforcement Fine-Tuning, this difficulty signal modulates the reward function. For simple queries characterized by high confidence and correctness, the model is penalized for redundancy. Conversely, for complex queries exhibiting high uncertainty or errors, the model is incentivized to extend its reasoning chain to explore alternative solutions.

\subsection{Uncertainty-Guided Difficulty Estimation}
A robust adaptive reasoning model must accurately distinguish between simple and complex queries. We propose that incorporating output-level uncertainty into the difficulty assessment is essential, as reliance solely on sampling accuracy overlooks the inherent variability of LLM generation.

\noindent\textbf{Output Uncertainty Calculation.}
Following prior work~\citep{wang2025beyond}, we quantify token-level uncertainty using Shannon entropy. For a generated sequence, the entropy $H_t$ at decoding step $t$ is defined as:
\begin{equation}
\label{eq:entropy}
\begin{split}
    H_t &:= - \sum_{j=1}^{V} p_{t,j} \log p_{t,j}, \\
    \text{s.t. } p_t &= \pi_\theta(\cdot \mid \mathbf{x}, s_{<t}) = \text{Softmax} \left( \frac{z_t}{T} \right),
\end{split}
\end{equation}
where $V$ is the vocabulary size, $z_t$ represents the logits, and $T$ is the temperature. We evaluate entropy using the training policy $\pi_\theta$ distribution. To represent the uncertainty of the entire trajectory $o_i$, we calculate the mean entropy of the top-$K$ most uncertain tokens, denoted as $\mathcal{H}_i$. Guided by \citep{wang2025beyond}, which suggests that the top 20\% of high-entropy tokens effectively represent sequence-level uncertainty, we adopt this ratio as our default setting for $K$. We further investigate the sensitivity of $K$ through ablation studies in Appendix~\ref{tab:ablation_k}. To ensure numerical stability across samples within a batch $\mathcal{B}$, we apply min-max normalization to these trajectory-level uncertainties:
\begin{equation}
\begin{split}
    \mathcal{H}_i &:= \frac{1}{K} \sum_{t \in \mathcal{T}_K} H_t, \\
    \widetilde{\mathcal{H}}_i &:= \frac{\mathcal{H}_{i} - \min_{j} \mathcal{H}_{j}}{\max_{j} \mathcal{H}_{j} - \min_{j} \mathcal{H}_{j}}, \\
    &\quad \forall i \in \{1, \dots, |\mathcal{B}|\times G\},
\end{split}
\end{equation}
where $\mathcal{T}_K$ is the set of indices corresponding to the top-$K$ entropy values in the sequence.

\paragraph{Difficulty Estimation.}
We define problem difficulty by integrating correctness with the normalized uncertainty. For a question with ground-truth $y^*$ and $G$ sampled rollouts, the difficulty score $\mathcal{D}_q$ is formulated as:
\begin{equation}
\label{eq:difficulty}
\mathcal{D}_q = 1 - \frac{1}{G} \sum_{i=1}^{G} \mathbb{I}[o_i = y^*] \cdot \Big( \alpha(1 - \widetilde{\mathcal{H}}_i) + (1-\alpha) \Big).
\end{equation}
Here, $\mathbb{I}[\cdot]$ is the indicator function and $\alpha$ balances the contribution of uncertainty. This formulation assigns higher difficulty scores to questions where the model is incorrect, or where it yields correct answers with high uncertainty.

\subsection{Difficulty-Aware Length Calibration}
To enforce adaptive reasoning, we modulate the reward based on the estimated difficulty relative to a dynamic batch threshold.

\paragraph{Dynamic Thresholding.}
We compute a difficulty threshold $\theta_{\mathcal{B}}$ as the $\tau$-quantile of difficulty scores within the batch $\mathcal{B}$. To ensure training stability, we apply an exponential moving average (EMA) update to the threshold and the average lengths for simple ($\bar{L}_s$) and hard ($\bar{L}_h$) questions:
\begin{equation}
    \theta_{\mathcal{B}} \leftarrow \gamma \cdot \theta_{\mathcal{B}}^{\text{cur}} + (1 - \gamma) \cdot \theta_{\mathcal{B}}^{\text{old}},
\end{equation}
where $\gamma$ is the momentum coefficient.

\paragraph{Length Reward.}
The length-based reward $\mathcal{R}_{\text{len}}^i$ is defined piecewise to penalize redundancy in easy tasks and encourage depth in hard tasks:
\begin{equation}
    \mathcal{R}_{\text{len}}^i = 
    \begin{cases} 
        \max\left(0, 1 - \dfrac{L_i}{\bar{L}_s} \cdot \rho_i \right), & \text{if } \mathcal{D}_q < \theta_{\mathcal{B}} \land o_i = y^*, \\[10pt]
        \min\left(1, \dfrac{L_i}{\bar{L}_h} \cdot \rho_i - 1 \right), & \text{if } \mathcal{D}_q > \theta_{\mathcal{B}} \land o_i \neq y^*, \\[10pt]
        0, & \text{otherwise}.
    \end{cases}
\end{equation}
where $L_i$ is the length of output $o_i$, and $\rho_i = 0.5 + 0.5(1-\widetilde{\mathcal{H}}_i)$ is a scaling factor dependent on uncertainty. The first condition discourages unnecessarily long outputs for simple, correctly answered questions. The second condition provides performance compensation by rewarding longer reasoning chains for difficult questions where the model initially fails. The remaining two quadrants---\emph{Hard+Correct} and \emph{Easy+Incorrect}---are deliberately assigned a zero length signal so that brevity is never penalized when a fragile reasoning chain happens to succeed, and length is never rewarded when the model has not yet produced a correct answer; in both regimes the accuracy reward $\mathcal{R}_{\text{acc}}$ dominates the gradient. Empirically, this asymmetric design is what prevents the collapse modes observed in greedy length-calibration baselines (Section~\ref{sec:results}); a fuller ablation is provided in Appendix~\ref{app:zero_reward_rationale}.

The total reward for each trajectory combines the accuracy reward $\mathcal{R}_{\text{acc}}$, format reward $\mathcal{R}_{\text{format}}$, and the adaptive length reward:
\begin{equation}
    \mathcal{R}_i = \mathcal{R}_{\text{acc}}^i + \lambda_1 \mathcal{R}_{\text{format}}^i + \lambda_2 \mathcal{R}_{\text{len}}^i.
\end{equation}
We employ default values $\lambda_1 = \lambda_2 = 0.5$. This composite reward is then utilized to optimize the policy via Group Relative Policy Optimization (GRPO).

\section{Experiments}
\noindent{\textbf{Training Datasets and Evaluation Benchmarks.}} We use the AlphaMed19k~\citep{liu2025beyond} dataset as our primary training corpus. AlphaMed19k integrates the official training splits \citep{huang2025m1} of two public multiple-choice medical QA benchmarks, MedQA \citep{medqa} and MedMCQA \citep{pal2022medmcqa}. MedQA contains expert-level clinical questions drawn from the USMLE, while MedMCQA covers factoid and reasoning questions from Indian medical entrance examinations. For out-of-domain evaluation we use PubMedQA \citep{jin2019pubmedqa}, the medical subsets of MMLU-Pro \citep{wang2024mmlu} (MMLU-ProM), the medical subsets of GPQA \citep{rein2023gpqa} (GPQA-M), and the recent clinical benchmark MedXpertQA \citep{zuo2025medxpertqa}. Additional dataset details are provided in the Appendix~\ref{appendix:dataset_details}.

\noindent{\textbf{Baseline \& Implementation Details.}} We compare AdaThink-Med with baseline models from three categories: general zero-shot LLMs (Qwen-2.5-Instruct-7B, Llama-3.1-Instruct-8B), medical tuned LLMs (MedLlama3, MMed, Med42~\citep{christophe2024med42}, OpenBioLLM~\citep{OpenBioLLMs}, UltraMedical~\citep{zhang2024ultramedical}, HuatuoGPT-o1~\citep{chen2024huatuogpt-o1}, m1~\citep{huang2025m1}), and length-calibration methods (Kimi1.5~\citep{team2025kimi}, ShortBetter~\citep{yi2025shorterbetter}, DAST~\citep{shen2025dast}, CosFn~\citep{yeo2025demystifying}). Because the $G$ rollouts required by our difficulty estimator are already produced by vanilla GRPO and entropy is computed from existing logits, AdaThink-Med adds less than 1\% wall-clock overhead per training step. More details of baselines and implementation are summarized in Appendix~\ref{sec_baseline_details} and Appendix~\ref{sec_supp_imple_details}, respectively. 

\noindent{\textbf{Evaluation Metrics.}} We report accuracy and output length to assess correctness against ground truth and token usage, respectively. To capture the trade-off between computational cost and performance, we adopt the \textbf{Accuracy-Efficiency Score (AES)}~\citep{yi2025shorterbetter, luo2025o1}, defined as:
\[
\text{AES} =
\begin{cases}
\alpha \cdot \Delta \text{Length} + \beta \cdot |\Delta \text{Acc}|, & \text{if } \Delta \text{Acc} \geq 0,\\
\alpha \cdot \Delta \text{Length} - \gamma \cdot |\Delta \text{Acc}|, & \text{if } \Delta \text{Acc} < 0.
\end{cases}
\]
Here, $\Delta \text{Length} = \frac{\text{Length}_{\text{baseline}} - \text{Length}_{\text{model}}}{\text{Length}_{\text{baseline}}}$ is the relative reduction in output length and $\Delta \text{Acc} = \frac{\text{Acc}_{\text{model}} - \text{Acc}_{\text{baseline}}}{\text{Acc}_{\text{baseline}}}$ is the relative change in accuracy. \textbf{We set $\alpha=1$, $\beta=3$, and $\gamma=5$ by default for fair comparison.} The asymmetric design places a larger penalty on accuracy drops than the reward for improvements, which aligns with practical preferences.

\subsection{Results}
\label{sec:results}
\noindent{\textbf{Performance Comparison with General and Medical Reasoning LLMs.}}
We evaluate AdaThink-Med against leading general and medical reasoning models to assess inference efficiency. As detailed in Table~\ref{tab:compare_w_medllm}, our approach achieves substantial efficiency gains across all six benchmarks. Most notably, AdaThink-Med realizes a 4.7$\times$ compression for the Qwen backbone (reducing average length from 497 to 106 tokens) with a slight accuracy improvement (+0.25\%), and a 6.4$\times$ compression for the Llama backbone (410 to 64 tokens) with minimal performance degradation (-1.13\%). Unlike baselines that rely on extensive datasets and heavy inference costs, AdaThink-Med establishes state-of-the-art efficiency using only 19k samples without distilled CoT data. Furthermore, the model exhibits clear adaptive behavior: it generates significantly shorter responses for straightforward tasks (e.g., MedQA) while allocating more computational resources to complex reasoning challenges (e.g., GPQA-M), validating the proposed adaptive thinking strategy.

\begin{table*}[t]
\centering
{
\caption{\textbf{Performance of models on in-domain (*) and out-of-domain medical QA benchmarks.} The best results are shown in \textbf{bold}. $+$ indicates reasoning ability obtained with a Chain-of-Thought prompt; $\dagger$ denotes supervised fine-tuning with long CoT data; $\diamond$ represents reinforcement learning; and $\ddagger$ refers to training with additional datasets. \colorbox{highlightrow}{Gray} rows indicate baseline reasoning models prior to adaptive optimization, while \colorbox{highlightrow_yellow}{yellow} rows denote the final \textbf{AdaThink-Med} adaptive models. For AES, the baseline is Llama-3.1-8B-Instruct$^+$.}
\label{tab:compare_w_medllm}
\resizebox{\textwidth}{!}{%
\begin{tabular}{llllllllllllllll}
\toprule
\multirow{2}{*}{Model} & \multicolumn{2}{c}{MedQA*} & \multicolumn{2}{c}{MedMCQA*} & \multicolumn{2}{c}{PubMedQA} & \multicolumn{2}{c}{MMLU-ProM} & \multicolumn{2}{c}{GPQA-M} & \multicolumn{2}{c}{MedXpert} & \multicolumn{2}{c}{Avg} & \multirow{2}{*}{AES} \\
\cmidrule(lr){2-3} \cmidrule(lr){4-5} \cmidrule(lr){6-7} \cmidrule(lr){8-9}
\cmidrule(lr){10-11} \cmidrule(lr){12-13} \cmidrule(lr){14-15}
& Acc. & Len. & Acc. & Len. & Acc. & Len. & Acc. & Len. & Acc. & Len. & Acc. & Len. & Acc. & Len. & \\
\midrule
\multicolumn{16}{c}{\itshape\textbf{Llama backbone}} \\[3pt]
Llama-3.1-8B-Instruct$+$  & 68.18 & 461 & 57.32 & 309 & {78.20} & 296 & 60.06 & 457 & 45.64 & 485 & 16.08 & 545 & 54.25 & 425 & {--} \\
Llama-3.1-8B-Instruct & 54.75 & 302 & 54.69 & 169 & 77.00 & {113} & 57.00 & 256 & 35.64 & 345 & 14.48 & 342 & 48.93 & 254 & -0.09 \\
MedLlama3-8B-v1$+$ & 50.90 & 187 & 41.47 & {153} & 46.00 & 133 & 30.68 & 234 & 32.05 & 265 & 12.20 & 310 & 35.55 & 214 & -1.23 \\
MedLlama3-8B-v2$+$ & 60.56 & 271 & 54.53 & 254 & 73.40 & 216 & 54.26 & 323 & 43.84 & 378 & 13.91 & 315 & 50.08 & 293 & -0.07 \\
MMed-8B$\dagger\ddagger+
$ & 49.01 & 888 & 43.82 & 843 & 59.40 & 832 & 34.65 & 856 & 31.79 & 831 & 13.06 & 938 & 38.62 & 865 & -2.47 \\
Med42-8B$\ddagger+$
 & 57.97 & 343 & 55.53 & 276 & 70.80 & 204 & 53.28 & 357 & 43.07 & 359 & 13.26 & 402 & 48.99 & 324 & -0.25 \\
OpenBioLLM-8B$\dagger\ddagger^\diamond$ & 50.51 & {183} & 44.84 & 160 & 55.40 & 141 & 38.82 & {153} & 31.28 & 167 & 12.16 & 207 & 38.84 & {169} & -0.82 \\
UltraMedical-8B-3$\dagger\ddagger^\diamond$ & 70.30 & 402 & 59.16 & 364 & 76.80 & 282 & 59.93 & 524 & {46.92} & 490 & 15.91 & 630 & 54.84 & 449 & -0.02 \\
UltraMedical-8B-3.1$\dagger\ddagger^\diamond$ & {74.46} & 486 & 62.65 & {450} & 78.50 & {392} & {62.93} & {603} & {46.92} & {588} & {16.89} & {710} & {57.06} & {538} & {-0.11} \\
HuatuoGPT-o1-8B$\dagger\ddagger^\diamond$ & \textbf{{76.35}} & {568} & \textbf{{62.82}} & {445} & \textbf{{79.80}} & {445} & {63.71} & 521 & \textbf{{54.35}} & 585 & {17.06} & 601 & \textbf{59.02} & {527} & {+0.02} \\
\rowcolor{highlightrow} 
GRPO-LLama$^\diamond$ & 72.50 & 438 & 61.60 & 299 & 78.50 & 290 & 63.51 & 442 & 46.92 & 461 & 17.26 & 529 & 56.72 & 410 & {+0.17} \\
\rowcolor{highlightrow_yellow} 
{\textbf{AdaThink-Med-LLama}} & 67.00 & \textbf{36} & 60.41 & \textbf{73} & 78.40 & \textbf{54 }& 63.58 & \textbf{85} & 47.94 & \textbf{100} & 16.20 & \textbf{35} & 55.59 & \textbf{64} & \textbf{{+0.92}} \\
\midrule
\multicolumn{16}{c}{\itshape\textbf{Qwen backbone}} \\[3pt]
Qwen2.5-7B-Instruct$+$ & 63.86 & {490} & 56.68 & 359 & 73.10 & 386 & 62.28 & 523 & 46.41 & {562} & 12.28 & {585} & 52.44 & {484} & - \\
Qwen2.5-7B-Instruct & 54.28 & 313 & 53.43 & 196 & 72.70 & 129 & 56.67 & 292 & 38.71 & 448 & 12.44 & 291 & 48.04 & 278 & -3.73 \\
{m1-7B$\dagger\ddagger
$} & 75.01 & {2161} & 62.32 & {1749} & 74.20 & {1080} &\textbf{{68.07}} & {2564} & 51.53 & {3887} & \textbf{{18.28}} & {3265} & 58.24 & {2451} & -4.54 \\
\rowcolor{highlightrow} 
GRPO-Qwen$^\diamond$ & 67.55 & 504 & 59.52 & 378 & 72.30 & 396 & {66.12} & 535 & 48.46 & 575 & 14.48 & 592 & 54.74 & 497 & +0.11 \\
\rowcolor{highlightrow_yellow} 
{\textbf{AdaThink-Med-Qwen}} & 68.34 & 79 & 58.74 & 109 & 73.50 & 62 & 66.45 & 121 & {48.72} & 175 & 14.16 & 88 & 54.99 & 106 & +0.93 \\
\bottomrule
\end{tabular}
}}
\end{table*}

\begin{figure*}[t]
    \centering
    \includegraphics[height=0.3\linewidth]{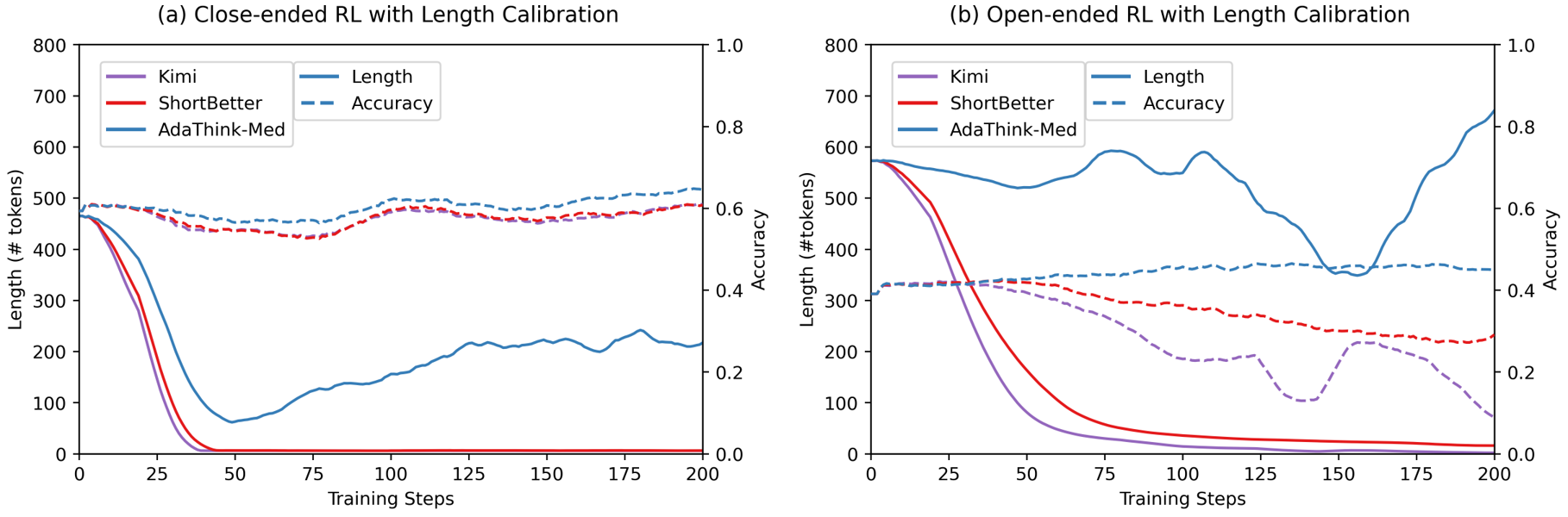}
    \caption{Length-reward hack during RL with greedy length calibration. Within each panel, accuracy (left $y$-axis) and average output length (right $y$-axis) are plotted against training step. \textbf{(a)} Closed-ended RL setup: both Kimi1.5 and ShortBetter collapse---accuracy and length drop sharply and do not recover, while AdaThink-Med (with performance compensation) preserves accuracy while compressing length. \textbf{(b)} Open-ended RL setup (following HuatuoGPT-o1): collapse persists even without candidate options, confirming that the failure is not an artifact of multiple-choice shortcuts. AdaThink-Med again maintains stable accuracy throughout training. For visual clarity, accuracy and length curves use distinct colors and markers.}
    \label{fig:len_reward_hack}
\end{figure*}

\noindent{\textbf{Length Reward Hack \& Motivation of AdaThink-Med's Performance Compensation.}} We conduct pilot experiments to investigate greedy output length reduction strategies that do not explicitly incentivize incorrect responses. Two representative approaches are Kimi1.5~\citep{team2025kimi}, which enforces uniform length reduction across all samples, and ShortBetter~\citep{yi2025shorterbetter}, which drives responses toward the minimal length among correct samples.
As shown in Fig.~\ref{fig:len_reward_hack}(a), the absence of such compensation causes both Kimi1.5 and ShortBetter to collapse in terms of output length and accuracy. We initially suspect that this collapse primarily arises in close-ended tasks, where the presence of candidate answers allows the model to adopt a trivial strategy of directly outputting the final answer without any reasoning. To avoid this shortcut, we adopt an open-ended RL setting following HuatuoGPT-o1~\citep{chen2024huatuogpt-o1}, where the model must reason and answer using its world knowledge without being provided candidate options~\citep{rui2025improving}. Implementation details are given in the Appendix. As shown in Fig.~\ref{fig:len_reward_hack}(b), even in the open-ended setting, greedy length calibration leads to collapse, with both accuracy and output length dropping sharply and failing to recover. Moreover, we found that
the collapsed model no longer produces meaningful answers or follows instructions. We attribute this phenomenon to a form of length reward hack: in its search for shorter responses, the model converges to a minimal-length strategy that maintains apparent accuracy while maximizing length-based rewards. This shortcut undermines the emergence of diverse reasoning patterns and degrades instruction following. To address this issue, we introduce a performance compensation mechanism that penalizes shorter outputs when accuracy degrades. 
This encourages the model to achieve a more balanced trade-off between performance and efficiency, ultimately leading to stable training dynamics as shown in Fig.~\ref{fig:len_reward_hack}.

\noindent\textbf{Comparison with Length-calibration Methods.}
As shown in Table~\ref{tab:compare_w_length_shap}, AdaThink-Med delivers marked efficiency gains over existing length-calibration baselines. It reduces the average reasoning length to 64 tokens on Llama and {106} tokens on Qwen. These lengths are {2.0}$\times$ and {2.8}$\times$ shorter than those produced by Kimi1.5, and 2.6$\times$ and {3.2}$\times$ shorter than those from ShortBetter, respectively, while achieving comparable accuracy (Llama: 55.59\% vs. 55.94\% for ShortBetter; Qwen: 54.99\% vs. 54.28\% for ShortBetter). Additionally, AdaThink-Med achieves the highest AES scores on both model backbones, with 0.92 on Llama and 0.93 on Qwen, respectively. This corresponds to an improvement of +0.14 over the next best method (Kimi1.5) on Llama and +0.38 on Qwen. These improvements stem from the model's adaptive thinking capability: AdaThink-Med dynamically allocates reasoning length, maintaining brevity on easy questions while employing extended reasoning chains on challenging or initially incorrect cases through its accuracy aware compensation mechanism. This approach effectively reduces computational cost without sacrificing reliability.
\begin{table*}[!t]
\centering
\caption{\textbf{Performance of length calibration methods on in-domain (*) and out-of-domain medical QA benchmarks.} For AES, baseline model is LLama-3.1-8B-Instruct$^+$ and Qwen-2.5-7B-Instruct$^+$, respectively.}
\label{tab:compare_w_length_shap}
\resizebox{\textwidth}{!}{%
\begin{tabular}{lccccccccccccccccr}
\toprule
\multirow{2}{*}{Model} & \multicolumn{2}{c}{MedQA*} & \multicolumn{2}{c}{MedMCQA*} & \multicolumn{2}{c}{PubMedQA} & \multicolumn{2}{c}{MMLU-ProM} & \multicolumn{2}{c}{GPQA-M} & \multicolumn{2}{c}{MedXpert} & \multicolumn{2}{c}{Avg} & \multirow{2}{*}{AES} \\
\cmidrule(lr){2-3} \cmidrule(lr){4-5} \cmidrule(lr){6-7} \cmidrule(lr){8-9}
\cmidrule(lr){10-11} \cmidrule(lr){12-13} \cmidrule(lr){14-15}
& Acc. & Len. & Acc. & Len. & Acc. & Len. & Acc. & Len. & Acc. & Len. & Acc. & Len. & Acc. & Len. & \\
\midrule
\multicolumn{16}{c}{\itshape\textbf{Llama backbone}} \\[3pt]
Llama-3.1-8B-Instruct$+$ & 68.18 & 461 & 57.32 & 309 & 78.20 & 296 & 60.06 & 457 & 45.64 & 485 & 16.08 & 545 & 54.25 & 425 & -- \\
Llama-3.1-8B-Instruct & 54.75 & 302 & 54.69 & 169 & 77.00 & 113 & 57.00 & 256 & 35.64 & 345 & 14.48 & 342 & 48.93 & 254 & -0.09 \\
\rowcolor{highlightrow} 
Llama-3.1-8B-Instruct-GRPO & \textbf{72.50} & 438 & \textbf{61.60} & 299 & 78.50 & 290 & 63.51 & 442 & 46.92 & 461 & \textbf{17.26} & 529 & \textbf{56.72} & 410 & +0.17 \\
Kimi1.5~\citep{team2025kimi} & 70.62 & 145 & 60.93 & 100 & 77.50 & 92 & \textbf{63.71} & 164 & 45.38 & 118 & 15.75 & 131 & 55.65 & 125 & +0.78 \\
CosFn~\citep{yeo2025demystifying} & 37.94 & 1007 & 49.29 & 919 & 75.10 & 885 & 23.32 & 998 & 32.82 & 1012 & 10.12 & 1020 & 38.10 & 974 & -2.78 \\
DAST~\citep{shen2025dast} & 60.64 &\textbf{30} &56.9 & \textbf{34}  & 76.00 & \textbf{39} & 59.48 &\textbf{71} & 	47.69&	\textbf{53}& 15.35& \textbf{82} & 52.71 & \textbf{51}& +0.74 \\
ShortBetter~\citep{yi2025shorterbetter} & 71.48 & 175 & 60.62 & 134 & \textbf{78.60} & 118 & 62.99 & 188 & 45.89 & 219 & 16.08 & 171 & 55.94 & 167 & +0.70 \\
\rowcolor{highlightrow_yellow} 
\textbf{AdaThink-Med-LLama} & 67.00 & 36 & 60.41 & 73 & 78.40 & 54 & 63.58 & 85 & \textbf{47.94} & 100 & 16.20 & 35 & 55.59 & 64 & \textbf{+0.92} \\ 
\toprule
\multicolumn{16}{c}{\itshape\textbf{Qwen backbone}} \\[3pt]
Qwen2.5-7B-Instruct$+$ & 63.86 & 490 & 56.68 & 359 & 73.10 & 386 & 62.28 & 523 & 46.41 & 562 & 12.28 & 585 & 52.44 & 484 & -- \\
Qwen2.5-7B-Instruct & 54.28 & 313 & 53.43 & 196 & 72.70 & 129 & 56.67 & 292 & 38.71 & 448 & 12.44 & 291 & 48.04 & 278 & +0.01 \\
\rowcolor{highlightrow} 
Qwen2.5-7B-Instruct-GRPO & 67.55 & 504 & 59.52 & 378 & 72.30 & 396 & 66.12 & 535 & 48.46 & 575 & 14.48 & 592 & 54.74 & 497 & +0.11 \\
Kimi1.5~\citep{team2025kimi} & \textbf{69.99} & 285 & 58.95 & 225 & 75.00 & 236 & \textbf{66.71} & 338 & 46.92 & 376 & 14.89 & 330 & \textbf{55.41} & 298 & +0.55 \\
CosFn~\citep{yeo2025demystifying} & 37.94 & 1007 & 49.29 & 919 & \textbf{75.10} & 885 & 23.32 & 998 & 32.82 & 1012 & 10.12 & 1020 & 38.10 & 974 & -2.38 \\
DAST~\citep{shen2025dast} & 65.90 & 260 & \textbf{59.67} & 327 & 69.70 & 153 & 65.86 & 205 & 50.51 & 466 & \textbf{15.59} & 288 & 54.54 & 283 & +0.54 \\
ShortBetter~\citep{yi2025shorterbetter} & 67.94 & 333 & 58.90 & 250 & 71.70 & 260 & 65.86 & 376 & 46.92 & 405 & 14.36 & 387 & 54.28 & 335 & +0.41 \\
\rowcolor{highlightrow_yellow} 
{\textbf{AdaThink-Med-Qwen}} & 68.34 & \textbf{79} & 58.74 & \textbf{109} & 73.50 & \textbf{62} & 66.45 & \textbf{121} & {48.72} & \textbf{175} & 14.16 & \textbf{88} & 54.99 & \textbf{106} & \textbf{{+0.93}} \\
\bottomrule
\end{tabular}
}
\end{table*}

\begin{figure*}[t]
    \centering\includegraphics[height=0.4\linewidth]{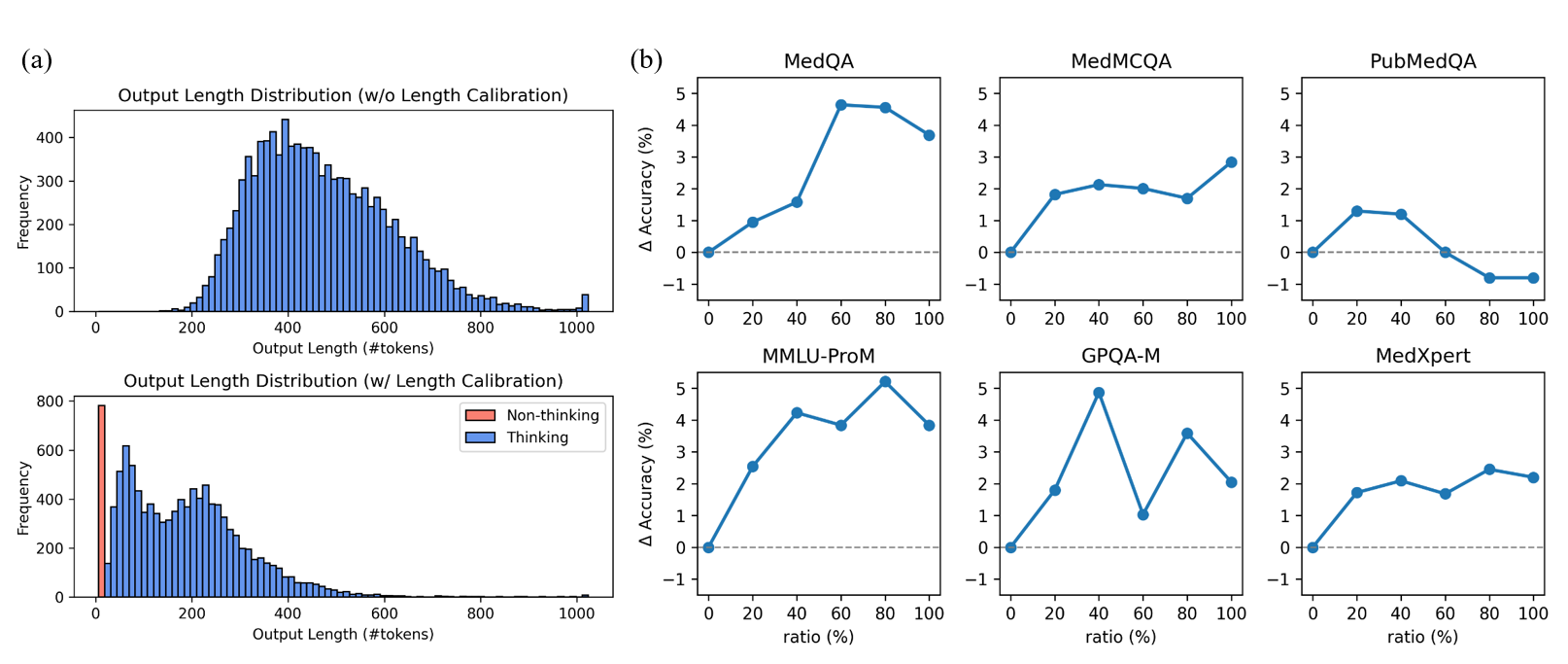}
    \caption{(a) Output length distribution w and w/o length calibration. (b) Dataset selection results.}
    \label{fig:len_distribution}
\end{figure*}

\noindent\textbf{Reasoning Mode Separation via Reward Design.}
As illustrated in Fig.~\ref{fig:len_distribution}(a), models trained with AdaThink-Med exhibit a notable bimodal distribution in reasoning length, separating into two behaviors: a \textit{non-thinking} mode that outputs answers directly without intermediate reasoning, and a \textit{thinking} mode characterized by concise yet essential reasoning steps. We do not claim this as a spontaneously emergent property---rather, it is a direct consequence of our difficulty-aware length reward, which jointly penalizes verbosity on easy-correct samples and rewards exploration on hard-incorrect ones. The contribution, instead, is that this mode separation arises \emph{end-to-end within a single model}, whereas prior work~\citep{zhang2025synapseroute} relies on manually annotated data and an external difficulty classifier to route between two separately trained models. To our knowledge, this is the first demonstration in medical QA that such bimodal reasoning behavior can be learned end-to-end without auxiliary supervision.

\noindent{\textbf{Dataset Selection Application.}} Dataset selection~\citep{albalak2024survey} aims to identify high-quality subsets that enable models to achieve comparable or superior performance with reduced training data. We leverage the intrinsic property of AdaThink-Med, where output length effectively correlates with problem difficulty, to filter high-value samples for reinforcement learning. Methodologically, we analyze the AlphaMed19k dataset using AdaThink-Med-Qwen to derive an output length distribution, subsequently partitioning samples into easy' (short) and hard' (long) categories based on the median length. We then construct stratified subsets at retention ratios of 20\%, 40\%, 60\%, and 80\% to investigate optimal data efficiency. As illustrated in Fig.~\ref{fig:len_distribution}(b), a 40\% retention ratio not only matches but exceeds full-dataset performance, yielding gains of 2.10\% on PubMedQA and 2.82\% on GPQA-M. Remarkably, even at 20\% retention, the model preserves 98.8\% of the full-set performance. Detailed control experiments demonstrating the superiority of this strategy over random subset selection are provided in the Appendix, confirming that our method selects intrinsically higher-quality training instances.

\begin{table*}[t]
\centering
\caption{Ablation study on $\tau$ and $\alpha$ across in-domain and out-of-domain datasets. We rewrite the result of $\tau=0.7, \alpha=0.5$ for clear comparison.}
\label{tab:ablation}
\resizebox{\textwidth}{!}{%
\begin{tabular}{ccccccccccccccccc}
\toprule
\multirow{2}{*}{$\tau$} & \multirow{2}{*}{$\alpha$} 
& \multicolumn{2}{c}{MedQA} & \multicolumn{2}{c}{MedMCQA} & \multicolumn{2}{c}{PubMedQA} 
& \multicolumn{2}{c}{MMLU-ProM} & \multicolumn{2}{c}{GPQA-M} & \multicolumn{2}{c}{MedXpert} 
& \multicolumn{2}{c}{Avg} & \multirow{2}{*}{AES} \\
\cmidrule(lr){3-4} \cmidrule(lr){5-6} \cmidrule(lr){7-8} \cmidrule(lr){9-10} 
\cmidrule(lr){11-12} \cmidrule(lr){13-14} \cmidrule(lr){15-16}
 & & Acc. & Len. & Acc. & Len. & Acc. & Len. & Acc. & Len. & Acc. & Len. & Acc. & Len. & Acc. & Len. & \\
\midrule
0.3 & 0.5 & 68.19 & 391 & 58.76 & 307 & 73.90 & 311  & 68.08 & 434 & 50 & 483 & 14.37 & 457 & 55.55 & 397 & 0.36 \\
0.5 & 0.5 & 67.32 & 105 & 59.12 & 170 & 72.80 & 84  & 67.17 & 196 & 47.69 & 245 & 15.02 & 116 & 54.85 & 153 & 0.82 \\
\rowcolor{highlightrow_yellow} 
0.7 & 0.5 & 68.34 & 79  & 58.74 & 109 & 73.50 & 62  & 66.45 & 121 & 48.72 & 175 & 14.16 & 88  & 54.99 & 106 & 0.93 \\
0.9 & 0.5 & 62.53 & 7   & 58.50 & 13  & 74.00 & 6   & 64.30 & 11  & 46.67 & 12  & 14.49 & 8   & 53.41 & 10  & 1.04 \\
\midrule
0.7 & 0.1 & 66.06 & 121 & 59.02 & 153 & 74.30 & 116 & 66.19 & 195 & 50.51 & 258 & 15.39 & 142 & 55.25 & 164 & 0.82 \\
\rowcolor{highlightrow_yellow} 
0.7 & 0.5 & 68.34 & 79  & 58.74 & 109 & 73.50 & 62  & 66.45 & 121 & 48.72 & 175 & 14.16 & 88  & 54.99 & 106 & 0.93 \\
0.7 & 0.9 & 63.47 & 44 & 58.12 & 216 & 73.60 & 10 & 63.19 & 50 & 46.92 & 69 & 14.08 & 30 & 53.23 & 70 & 0.90 \\
\bottomrule
\end{tabular}%
}
\end{table*}

\noindent\textbf{Human Evaluation: Clinical Validity}
To assess whether the efficiency gains compromise clinical rigor, we conducted a blind human evaluation involving three board-certified Internal Medicine physicians (avg.~12 years of clinical experience; Fleiss' $\kappa=0.76$, indicating substantial agreement; disagreements resolved via consensus) on 500 randomly sampled test cases, measuring \textit{Response Accuracy}, \textit{Reasoning Sufficiency} (the retention of essential logical steps), and \textit{Logical Soundness} (reasoning chains free from hallucinations or flawed deductions). As detailed in Table~\ref{tab:human_eval}, aggressive length-reduction baselines exhibit ``shortcut hallucinations''---producing concise but medically flawed rationales to exploit brevity rewards, resulting in Logical Soundness below 70\%. In contrast, AdaThink-Med achieves the best results on all three axes (Accuracy 76.25\%, Sufficiency 89.00\%, Logical Soundness 86.40\%), demonstrating that our uncertainty-guided mechanism only compresses reasoning when the model is both correct and confident, preserving the fidelity of clinical decision-making.

\begin{table}[!t]
\centering
\caption{Human evaluation results on 500 randomly sampled test cases by three board-certified Internal Medicine physicians (Fleiss' $\kappa=0.76$). \textbf{Reasoning Sufficiency} measures retention of essential logical steps; \textbf{Logical Soundness} measures whether the reasoning chain is free from hallucinations or flawed deductions. AdaThink-Med achieves the best balance across all three axes.}
\label{tab:human_eval}
\resizebox{\linewidth}{!}{%
\begin{tabular}{lccc}
\toprule
\textbf{Method} & \textbf{Accuracy (\%)} & \textbf{Reasoning Sufficiency (\%)} & \textbf{Logical Soundness (\%)} \\
\midrule
Kimi            & 71.25 & 73.00 & 68.40 \\
CosFn           & 70.34 & 67.00 & 65.20 \\
DAST            & 68.37 & 65.00 & 62.80 \\
ShortBetter     & 70.23 & 72.00 & 69.50 \\
\textbf{AdaThink-Med}   & \textbf{76.25} & \textbf{89.00} & \textbf{86.40} \\
\bottomrule
\end{tabular}%
}
\end{table}

\noindent\textbf{Real-World Open-Ended Clinical Evaluation.}
To validate clinical utility beyond multiple-choice QA, we additionally conducted an open-ended evaluation on \textbf{796 anonymized real-world cardiac stroke cases} curated in collaboration with four hospitals. Each model receives the patient's clinical record and must generate a free-text \emph{Diagnosis} and a free-text \emph{Treatment Plan}; three board-certified clinicians then blindly score every generation on \emph{Accuracy}, \emph{Redundancy} (lower is better), and \emph{Clinical ICA} (logical soundness and completeness). As shown in Table~\ref{tab:clinical_open_ended}, AdaThink-Med attains the best Accuracy and Clinical ICA on both Diagnosis and Treatment, while keeping Redundancy at 18.6\%/24.3\%. Greedy length-reduction baselines (Kimi, ShortBetter) further reduce redundancy but at the cost of Clinical ICA dropping below 70\%, indicating systematic over-pruning of clinically essential reasoning. This directly evidences, in a clinically realistic free-text generation setting, that our uncertainty-guided mechanism preserves diagnostic rigor while removing genuinely redundant content.

\begin{table}[!t]
\centering
\caption{\textbf{Open-ended evaluation on 796 real-world cardiac stroke cases}, scored blindly by three board-certified clinicians. \emph{Diag.}~$=$~Diagnosis; \emph{Treat.}~$=$~Treatment Plan; \emph{Red.}~$\downarrow$~lower is better; \emph{ICA}~$=$~Clinical logical soundness and completeness.}
\label{tab:clinical_open_ended}
\resizebox{\linewidth}{!}{%
\begin{tabular}{lcccccc}
\toprule
\textbf{Method} & \textbf{Diag.\ Acc.} & \textbf{Diag.\ Red.\ $\downarrow$} & \textbf{Diag.\ ICA} & \textbf{Treat.\ Acc.} & \textbf{Treat.\ Red.\ $\downarrow$} & \textbf{Treat.\ ICA} \\
\midrule
Standard GRPO   & 81.7 & 68.7 & 76.3 & 74.5 & 75.1 & 72.9 \\
Kimi            & 78.3 & 35.2 & 71.3 & 70.5 & 38.4 & 64.2 \\
CosFn           & 77.2 & 42.1 & 68.5 & 69.2 & 45.3 & 62.5 \\
DAST            & 76.6 & 38.2 & 66.7 & 68.2 & 40.8 & 60.7 \\
ShortBetter     & 77.2 & 25.4 & 69.4 & 68.7 & 22.1 & 61.5 \\
\textbf{AdaThink-Med} & \textbf{83.2} & \textbf{18.6} & \textbf{78.5} & \textbf{77.8} & \textbf{24.3} & \textbf{74.2} \\
\bottomrule
\end{tabular}}
\end{table}

\subsection{Ablations}
\noindent{\textbf{Impact of Problem Difficulty $\tau$ and Uncertainty Weight $\alpha$.}} Since the parameter $\tau$ coarsely determines the overall compression ratio of the response length, we first fix the difficulty metric by setting the weight $\alpha$ for the uncertainty measure to 0.5, in order to investigate the impact of the difficulty threshold $\tau$. A  grid-search over $\{0.3, 0.5, 0.7, 0.9\}$ is conducted for the optimal $\tau$. As shown in Table~\ref{tab:ablation}, when $\tau$ is set too high (0.9), the average length on the test set is substantially reduced along with a degradation in performance and length reward hack. Conversely, $\tau=0.3$ produces longer reasoning chains across the dataset (redundant over-thinking), even though performance improves slightly. 
Next, we examine two values for $\alpha$, i.e., 0.1 and 0.9, while fixing $\tau$. Table~\ref{tab:ablation} shows that $\alpha=0.1$ (accuracy-focused) gives weaker performance (AES 0.82), while $\alpha=0.9$ (uncertainty-focused) leads to only a slight drop (AES 0.90), highlighting the benefit of uncertainty-based difficulty for efficient reasoning.
$\alpha=0.5$ exhibits the best performance among the candidates. We believe an over-reliance on the uncertainty measure ($\alpha=0.9$) may lead to a lag in performance compensation as its estimation precision increases. Overall, $\alpha=0.5$ is identified as the optimal setting under the balanced $\tau=0.7$.
\begin{figure*}[t]
    \centering
    \includegraphics[width=0.90\linewidth]{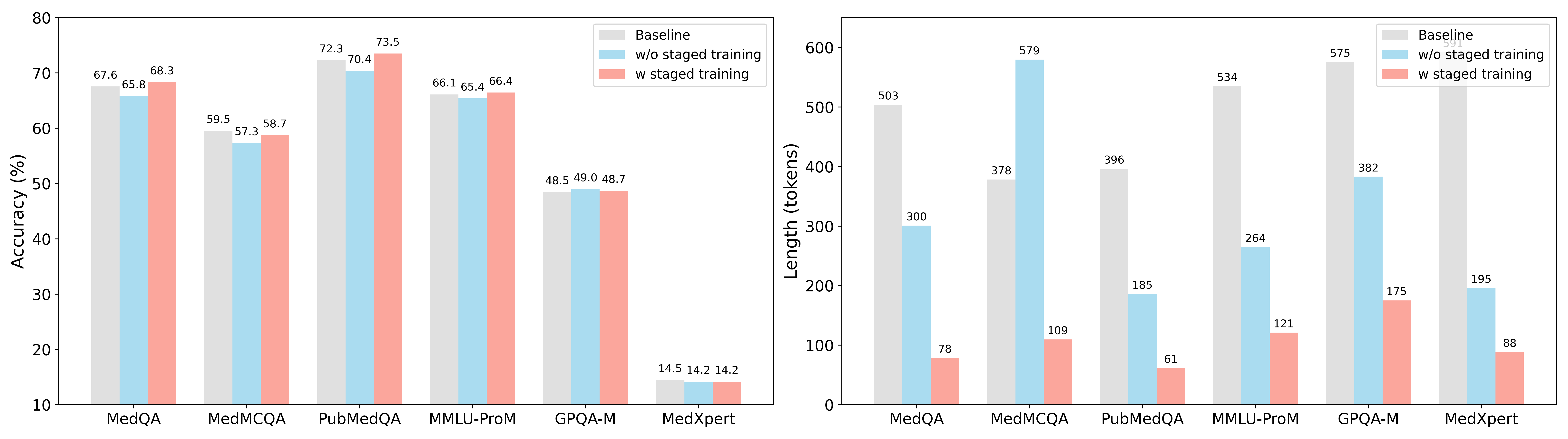}
    \caption{Performance comparison w and w/o staged adaptive training.}
    \label{fig:staged_training} 
\end{figure*}

\noindent\textbf{Effectiveness of Staged Adaptive Training.} In the initial phase of reinforcement learning, the model often exhibits low prediction accuracy and significant uncertainty, causing most questions to be perceived as hard. Emphasizing samples with incorrect outputs, our model inherently prioritizes accuracy improvement, leading to an imbalance between reducing overthinking and enhancing overall performance.
To address this, we introduce a staged adaptive training strategy: the base model is first trained to convergence without length calibration, after which adaptive training is applied. 
As illustrated in Fig.~\ref{fig:staged_training}, omitting the staged approach leads to limited length compression and noticeable performance degradation. In contrast, staged training maintains performance while achieving greater reduction in output length, demonstrating its effectiveness.

\section{Conclusion}
In this paper, we propose \textbf{AdaThink-Med}, an Uncertainty-Guided adaptive thinking framework designed to improve the efficiency of medical large language models. By dynamically shaping output lengths based on problem difficulty, AdaThink-Med effectively reduces redundant reasoning while maintaining competitive accuracy. Furthermore, we propose a data selection strategy that identifies high-quality reasoning and non-reasoning samples, enabling efficient training with significantly fewer data. Extensive experiments across six medical reasoning benchmarks demonstrate that AdaThink-Med achieves state-of-the-art overall performance under the trade-off between efficiency and accuracy, and produces a clear ``thinking'' vs.\ ``non-thinking'' mode separation within a single model as a direct consequence of our reward design, eliminating the need for external routers or classifiers used by prior pipeline approaches. We further discuss key limitations and promising future directions in Appendix~\ref{app:limitations}.

\section*{Limitations and Future Work}
\label{app:limitations}
While AdaThink-Med achieves strong empirical efficiency gains, several limitations should be acknowledged. (i) \textbf{Reliance on entropy quality.} Our difficulty estimator depends on output entropy derived from model-generated rollouts; if the model is severely miscalibrated or rollouts are highly homogeneous, entropy may become a weak proxy for true difficulty. Although we observed that intra-batch min-max normalization and EMA smoothing already make our pipeline robust in practice (Appendix~\ref{appendix:entropy_sensitivity}), more principled uncertainty estimators (e.g., learned calibration heads or process reward models) deserve further investigation.
(ii) \textbf{Step-level verification.} The current framework operates at the sequence level. Integrating step-level verifiers such as Process Reward Models (PRMs) could provide finer-grained signals on which reasoning step is redundant versus essential, potentially improving Logical Soundness further.
(iii) \textbf{Single-turn QA scope.} Our evaluation, while spanning six benchmarks and a real-world cardiac stroke study (Appendix~\ref{app:clinical_open_ended}), is primarily single-turn. In multi-turn clinical workflows (e.g., iterative differential diagnosis, longitudinal consultations) or in long-context settings, reasoning traces accumulate over multiple turns and significantly degrade both performance and cost; the ability of AdaThink-Med to keep reasoning concise on simple turns is expected to be particularly beneficial in such regimes, and represents a natural direction for future deployment studies.
(iv) \textbf{Cross-domain transfer.} Although the core mechanism is domain-agnostic, calibration of $\tau$ and $\alpha$ may need to be re-tuned when transferring across specialties, languages, or patient populations. Systematic study of cross-domain transfer of difficulty thresholds is left to future work.
(v) \textbf{Backbone-specific stabilization.} As detailed in Appendix~\ref{app:rep_pen_ablation}, Llama backbones additionally require an $n$-gram repetition penalty to prevent reward hacking via degenerate repetition; this fragility motivates further work on backbone-agnostic stabilization techniques.

\section*{Acknowledgments}
This work was supported by the Shanghai Innovation Institute.

\section*{Impact Statement}
By dynamically optimizing inference compute based on clinical complexity, AdaThink-Med reduces the latency, cost, and energy footprint of medical reasoning models in clinical workflows. We emphasize, however, that AdaThink-Med is a clinical \emph{decision-support} tool requiring rigorous prospective validation and human-in-the-loop oversight before deployment; calibration may also need to be re-validated when transferring across patient populations, languages, or specialties.

\bibliography{ref}
\bibliographystyle{icml2026}

\clearpage
\onecolumn
\appendix
\startcontents[appendix]
\section*{Appendix Contents}
\printcontents[appendix]{l}{1}{\setcounter{tocdepth}{2}}
\clearpage

\section{Reproducibility Statement}
We place strong emphasis on reproducibility in this work. Detailed descriptions of experimental configurations—including model architectures, hyperparameter settings, and training strategies—are provided in Appendix~\ref{sec_baseline_details}, with further implementation specifics available in Appendix~\ref{sec_supp_imple_details}. To further support transparent validation, the complete source code and step-by-step instructions for replicating all experiments are publicly available at \url{https://github.com/shaohao011/AdaThinkMed}.

\section{More Related Works}
\label{app:more_related_works}
{\noindent\textbf{Medical LLMs.}
With rapid advancements in training data and synthetic data formulation, LLMs in the medical domain have achieved significant performance improvements. Notable developments include instruction-tuned medical chatbot series~\citep{OpenBioLLMs,qiu2024towards,christophe2024med42,zhang2024ultramedical}, reasoning models incorporating Chain-of-Thought (CoT) methodologies~\citep{chen2024huatuogpt-o1,huang2025m1,rui2026cardiocot}, and agent collaboration~\citep{chen2025mediator}. These models have seen substantial adoption in both academic research and real-world applications. However, for clinical deployment, computational efficiency remains a critical concern alongside diagnostic accuracy. While SynapseRoute~\citep{zhang2025synapseroute} explores a router network to partition LLMs into ``thinking" and ``non-thinking" pathways, this approach requires additional manual labeling to train difficulty-classification networks, incurring extra computational and human resource costs. In contrast, we propose an efficient method that allocates computational resources within a single model without requiring supplementary modules or human intervention, thereby expanding potential application domains.}

\noindent\textbf{Difficulty-Aware Length Control.}
A concurrent line of work directly targets difficulty-aware reasoning length control. Selective Chain-of-Thought for medical QA~\citep{zhan2026reason} introduces an explicit routing classifier that decides at inference whether to invoke a long CoT, requiring a separately trained router. L1~\citep{aggarwal2025l1} learns a length controller jointly with the policy via reinforcement learning, but uses a static correctness signal and does not model predictive uncertainty. The Overthinker's DIET~\citep{chen2026overthinker} uses static problem-difficulty estimates obtained offline and applies fixed length budgets per difficulty bucket. Compared with these methods, AdaThink-Med (i) operates end-to-end within a single model with \emph{no} external router or classifier; (ii) couples \emph{dynamic} rollout correctness with \emph{predictive uncertainty} (Top-$K$ token entropy) inside the RL reward, making the difficulty signal co-evolve with the policy; and (iii) is tailored to the extreme complexity variance of clinical queries, where length budgets cannot be statically pre-assigned per question.

\noindent\textbf{Detailed Differentiation from Entropy-Guided Methods.}
We further detail the differences between AdaThink-Med and three representative entropy-based reasoning techniques.
(1) \emph{vs.\ SEED-GRPO}~\citep{chen2025seed}: SEED-GRPO uses semantic entropy to modulate the magnitude of policy updates (smaller gradients for uncertain queries) but leaves the reward function unchanged. AdaThink-Med fuses Top-$K$ token entropy with rollout correctness to \emph{redesign the RL reward}, directly teaching the model to adjust generation length based on difficulty.
(2) \emph{vs.\ Adaptive Termination}~\citep{xu2025adaptive}: This is a strict test-time intervention that externally halts parallel reasoning branches once a semantic entropy threshold is crossed. AdaThink-Med is a \emph{training-time} solution---length planning is internalized into the model's weights via RL, enabling autonomous mode switching in a single pass without an external controller.
(3) \emph{vs.\ Step-Entropy Compression}~\citep{li2025compressing}: This inserts \texttt{[SKIP]} tokens to prune low-entropy reasoning steps; rigid pruning can break reasoning on hard problems. AdaThink-Med operates at the \emph{sequence level}, fusing uncertainty with correctness into a Difficulty Score that provides ``performance compensation''---compressing easy queries while \emph{incentivizing extended reasoning} on hard cases, avoiding accuracy collapse from rigid pruning.

\section{Problem Formulation}
\label{app:preliminaries}
Efficient reasoning aims to optimize a reasoning model $\pi_\theta$ parameterized by $\theta$ 
to generate high-quality outputs with reduced inference cost. 
Given a dataset $\mathcal{D}$, where each sample consists of an input prompt $\mathbf{x}$ and the corresponding ground-truth answer $\mathbf{y}^*$, 
the model produces an output $\mathbf{y} \sim \pi_\theta(\cdot|\mathbf{x})$. A reward function $R(\mathbf{y}, \mathbf{y}^*)$ is defined to evaluate the correctness of the generated output $\mathbf{y}$. 
To ensure stable optimization and prevent overfitting, a reference model $\pi_{\mathrm{ref}}$ 
is introduced to regularize the policy update via a KL-divergence penalty term. 
The optimization objective for efficient reasoning is defined as:
\begin{equation}
\label{eq:object}
\max_{\theta} \mathbb{E}_{\mathbf{x} \sim \mathcal{D}, \mathbf{y} \sim \pi_\theta(\cdot|\mathbf{x})} \Big[ R(\mathbf{y}, \mathbf{y}^*) - \lambda \cdot \mathcal{C}(\pi_\theta, \mathbf{x}) - \beta \cdot D_{\mathrm{KL}}\big(\pi_\theta(\mathbf{y}|\mathbf{x}) \parallel \pi_{\mathrm{ref}}(\mathbf{y}|\mathbf{x})\big) \Big]
\end{equation}
where $\mathcal{C}(\pi_\theta, \mathbf{x})$ denotes the computational cost of generating $\mathbf{y}$ for the input $\mathbf{x}$, 
$\lambda$ controls the trade-off between reward and computational efficiency, 
and $\beta$ controls the influence of the KL-divergence regularization term. 
The reference model $\pi_{\mathrm{ref}}$ corresponds to the model state prior to efficient reasoning optimization. We use GRPO as our reinforcement learning method and the details can be found at Appendix~\ref{sec_app:exp_setup}.

\section{RL Algorithm}
\label{sec_app:exp_setup}
\noindent\textbf{Group Relative Policy Optimization (GRPO).} 
To optimize the efficient reasoning objective in Eq.~\eqref{eq:object}, 
both critic-based reinforcement learning methods (e.g., PPO) and critic-free methods (e.g., GRPO~\citep{guo2024deepseek}) can be applied. 
Considering its simplicity and effectiveness, GRPO has been widely adopted in recent works aiming to enhance the reasoning ability of large language models. Given an input prompt $\mathbf{x}$, a group of $G$ candidate outputs 
$\{o_i\}_{i=1}^G$ is sampled from the current policy $\pi_\theta$, 
where each $o_i = \{o_{i,1}, \dots, o_{i,|o_i|}\}$ corresponds to one possible output sequence $\mathbf{y}$.
We reformulate the objective of maximizing the expected reward $R(\mathbf{y}, \mathbf{y}^*)$ in Eq.~\eqref{eq:object} 
as minimizing the following GRPO loss:
\begin{equation}
\label{eq:grpo}
\begin{split}
\mathcal{L}_{\mathrm{GRPO}}(\theta) &= - \frac{1}{\sum_{i=1}^{G} |o_i|} \sum_{i=1}^{G} \sum_{t=1}^{|o_i|} \min \Bigg( \\
&\quad \frac{\pi_\theta(o_{i,t} \mid \mathbf{x}, o_{i,<t})}{\pi_{\theta_{\mathrm{old}}}(o_{i,t} \mid \mathbf{x}, o_{i,<t})}\hat{A}_{i,t}, \\
&\quad \text{clip}\left(
\frac{\pi_\theta(o_{i,t} \mid \mathbf{x}, o_{i,<t})}{\pi_{\theta_{\mathrm{old}}}(o_{i,t} \mid \mathbf{x}, o_{i,<t})},
1 - \epsilon, 
1 + \epsilon
\right)\hat{A}_{i,t}
\Bigg).
\end{split}
\end{equation}
where $\hat{A}_{i,t}$ denotes the relative advantage of token $o_{i,t}$ within the group, 
$\pi_{\theta_{\mathrm{old}}}$ represents the policy before the update.

\section{AdaThink-Med Evaluation Details}

\subsection{Baseline Models}
\label{sec_baseline_details}
We compare AdaThink-Med with strong large language models from both general-purpose and medical domains to enable a comprehensive evaluation. The general-purpose baselines are Qwen-2.5-Instruct-7B and Llama-3.1-Instruct-8B. The medical-domain baselines include MedLlama3, MMed, Med42~\citep{christophe2024med42}, OpenBioLLM~\citep{OpenBioLLMs}, UltraMedical~\citep{zhang2024ultramedical}, HuatuoGPT-o1~\citep{chen2024huatuogpt-o1}, and m1~\citep{huang2025m1}. Length calibration RL based baselines include representative Kimi1.5~\citep{team2025kimi}, ShortBetter~\citep{yi2025shorterbetter},DAST~\citep{shen2025dast}and CosFn~\citep{yeo2025demystifying}. To counteract the length reward hack issue observed in models like Kimi1.5 and ShortBetter, we select the checkpoint from the stable performance period preceding the point where accuracy declines by more than 10\% from its peak.

\subsection{Implementation Details}
\label{sec_supp_imple_details}
\noindent{\textbf{Basic Setups.}} Training is conducted using 8$\times$H100 GPUs (80GB VRAM each) with PyTorch, leveraging FlashAttention-2 for computational efficiency. GRPO-related experiments are implemented using the \texttt{verl}~\citep{sheng2024hybridflow} framework to accelerate the training. We adopt the Qwen2.5~\citep{qwen2.5}-7B-Instruct model and Llama3.1-Instruct-8B as the backbone models. The training uses a total batch size of 256, with a constant learning rate of $1\text{e}^{-6}$. The KL penalty coefficient is set to $\beta = 0.01$. For inference during GRPO, we deploy the model using \texttt{vllm}~\citep{kwon2023efficient} on 2 GPUs, generating 8 completions per sample, corresponding to the group size $G$ in GRPO. The temperature for \texttt{vllm} sampling is set to 1.0. As for our staged adaptive training, we first train the model using GRPO for 300 steps until full convergence is achieved. Subsequently, the proposed length calibration method is applied and train for 200 steps.

\noindent{\textbf{Computational Overhead of the Difficulty Estimator.}} The multiple rollouts required by our difficulty estimator are \emph{inherent} to GRPO (which, being critic-free, already samples $G{=}8$ rollouts per prompt to compute relative advantages); unlike PPO, no separate critic network is needed. Our additional cost is therefore limited to (i) reading already-materialized token logits, (ii) computing Top-$K$ Shannon entropy, and (iii) a batch-level min-max normalization plus EMA threshold update. In wall-clock terms this adds \textbf{less than 1\% overhead} per step on top of vanilla GRPO, since no extra forward passes are needed.

\noindent{\textbf{Repetition Penalty with N-gram on LLama- Models}}. We also observe that the model exhibits length reward hacking, wherein it increases the length of reasoning chains on incorrectly answered hard questions through repetition rather than substantive reasoning. This finding aligns with the observations in \citet{yeo2025demystifying}. However, such length reward hacking is not observed in Qwen backbones, a discrepancy that warrants further investigation. In our experiments with LLaMA backbones, we apply the same n-gram repetition penalty as described in their work.

\subsection{Dataset Details}
\label{appendix:dataset_details}
We conduct experiments on six publicly accessible medical and biomedical benchmarks, covering different knowledge domains and levels of reasoning difficulty. The datasets are summarized as follows:

\noindent\textbf{MedQA~\citep{medqa}.} This benchmark is derived from the United States medical licensing examination and is designed to test comprehensive clinical knowledge across a wide range of medical specialties. The standard test split is used for evaluation.

\noindent\textbf{MedMCQA~\citep{pal2022medmcqa}.} Originating from Indian medical entrance examinations, this dataset consists of multiple-choice questions that focus on fundamental medical concepts. Evaluation is performed on the official test set provided by the dataset.

\noindent\textbf{PubMedQA~\citep{jin2019pubmedqa}.} A biomedical question answering dataset constructed from PubMed abstracts. Each sample requires choosing between ``yes,'' ``no,'' or ``maybe'' to reflect factual understanding in biomedical research literature. We adopt the official test partition for all experiments.

\noindent\textbf{MMLU-ProM~\citep{wang2024mmlu}.} This is the medical subset of the general MMLU benchmark, specifically targeting professional-level medical and health-related knowledge. Following prior work~\citep{chen2024huatuogpt-o1}, we employ the standard split configuration defined for this subset.

\noindent\textbf{GPQA-M~\citep{rein2023gpqa}.} This dataset represents the biomedical branch of the Graduate-Level Question Answering benchmark. Its questions are manually curated to avoid superficial retrieval and to encourage deeper reasoning. We report results using the official evaluation split.

\noindent\textbf{MedXpert~\citep{zuo2025medxpertqa}.} A challenging benchmark built to assess advanced medical knowledge, complex reasoning, and clinical decision-making skills. It contains board-style exam questions across multiple body systems and specialties, curated by domain experts to ensure reliability and difficulty.

\section{Additional Experimental Results}

\subsection{Design Rationale: Zero-Reward Quadrants}
\label{app:zero_reward_rationale}
The piecewise length reward (Eq.~5) intentionally assigns \emph{zero} length signal to two quadrants: \textsc{Easy+Incorrect} ($\mathcal{D}_q<\theta_{\mathcal{B}}\wedge o_i\neq y^*$) and \textsc{Hard+Correct} ($\mathcal{D}_q>\theta_{\mathcal{B}}\wedge o_i=y^*$). This is a deliberate design choice that prioritizes accuracy over efficiency, justified by two observations from our early pilots with a symmetric 4-case reward:
\begin{itemize}[leftmargin=*]
\item \textbf{Hard+Correct.} Applying a brevity penalty in this quadrant caused aggressive over-pruning of essential diagnostic reasoning chains and catastrophic accuracy drops on complex datasets such as GPQA-M. Zeroing the brevity penalty here \emph{protects} fragile but successful reasoning chains, allowing the policy to consolidate them before being asked to compress.
\item \textbf{Easy+Incorrect.} Penalizing length here discourages self-correction; rewarding length encourages hallucinated padding. Zeroing the length signal lets the missed accuracy reward $\mathcal{R}_{\text{acc}}{=}0$ dominate the gradient, forcing the model to first \emph{fix the factual error} before any length-shaping is applied.
\end{itemize}
Empirically, this asymmetric design is what enables AdaThink-Med to compress aggressively (4.7--6.4$\times$) \emph{without} the collapse modes observed in greedy length-calibration baselines (Kimi1.5, ShortBetter).

\subsection{Empirical Distribution of Difficulty Quadrants}
\label{app:four_regime}
We further characterize the four (correctness, uncertainty) quadrants induced by Eq.~3 to clarify how often each regime is encountered during training and to address the concern that ``correct-but-uncertain'' cases (\emph{Lucky Correct}) might be misclassified as hard.
Table~\ref{tab:four_regime} reports the empirical proportions during the initial phase of adaptive RL on AlphaMed19k.
\begin{table}[h]
\centering
\caption{Empirical distribution of the four (correctness, uncertainty) quadrants at the start of adaptive RL. The ``Lucky Correct'' regime represents only $\sim$14\% of samples and progressively shrinks to $\sim$6\% as training stabilizes.}
\label{tab:four_regime}
\begin{tabular}{lccc}
\toprule
\textbf{Regime} & \textbf{Correctness} & \textbf{Uncertainty} & \textbf{Proportion} \\
\midrule
True Easy        & Correct   & Low  & 36.3\% \\
True Hard        & Incorrect & High & 30.8\% \\
Confident Error  & Incorrect & Low  & 18.5\% \\
Lucky Correct    & Correct   & High & 14.4\% \\
\bottomrule
\end{tabular}
\end{table}
The \textsc{Lucky Correct} regime accounts for only 14.4\% initially and shrinks to $\sim$6\% as training progresses. Crucially, in ablations that \emph{do} penalize length in this regime, the model learns truncated unreasoned guesses, degrading accuracy on related test questions. Withholding the length penalty allows the necessary token budget for the model to build rigorous evidence chains, gradually transitioning ``lucky guesses'' into ``confident deductions.''

\subsection{Per-Benchmark Difficulty Scores}
\label{app:per_benchmark_dq}
Although $\theta_{\mathcal{B}}$ is computed globally over the training distribution (AlphaMed19k), we verify that the induced difficulty estimator generalizes meaningfully to out-of-domain (OOD) benchmarks. Table~\ref{tab:per_benchmark_dq} reports per-benchmark average difficulty scores $\bar{\mathcal{D}}_q$ obtained by running the trained AdaThink-Med-Qwen offline on each test set.
\begin{table}[h]
\centering
\caption{Per-benchmark average difficulty scores $\bar{\mathcal{D}}_q$ from AdaThink-Med-Qwen. The model correctly ranks PubMedQA as easiest and MedXpert/GPQA-M as hardest, even for OOD benchmarks unseen during training.}
\label{tab:per_benchmark_dq}
\begin{tabular}{lcc}
\toprule
\textbf{Benchmark} & \textbf{Split} & \textbf{Avg.\ $\bar{\mathcal{D}}_q$} \\
\midrule
PubMedQA    & OOD       & 0.27 \\
MedQA       & In-domain & 0.31 \\
MMLU-ProM   & OOD       & 0.46 \\
MedMCQA     & In-domain & 0.52 \\
GPQA-M      & OOD       & 0.63 \\
MedXpert    & OOD       & 0.79 \\
\bottomrule
\end{tabular}
\end{table}
The ordering matches both expert intuition and downstream accuracy: PubMedQA (factoid yes/no/maybe) is identified as the easiest, while GPQA-M and MedXpert---which contain expert-level clinical reasoning---are correctly identified as the hardest. This indicates that a single global threshold trained on AlphaMed19k transfers to OOD difficulty estimation without per-benchmark re-tuning.

\subsection{Dynamic Threshold Evolution During Training}
\label{app:threshold_evolution}
Because $\theta_{\mathcal{B}}$ is the $\tau$-quantile of intra-batch difficulty scores under EMA smoothing, it acts as a \emph{relative} (rather than absolute) boundary that adapts as the model improves. Empirically we observe a two-phase trajectory: in the first half of adaptive RL, the threshold exhibits a \emph{downward drift}---as accuracy improves and entropy drops on previously hard examples, batch-wise $\mathcal{D}_q$ values decrease, lowering the quantile. In the second half, after the policy stabilizes, the threshold \emph{converges} to a stable value. The EMA coefficient $\gamma$ damps batch-to-batch variance and prevents abrupt reward flips during this drift. This dynamic relative behavior is precisely what allows the estimator to keep tracking the model's evolving capability rather than freezing to an outdated notion of difficulty.

\subsection{Extended Ablation on the Uncertainty Weight $\alpha$}
\label{app:alpha_extended}
For completeness we extend the $\alpha$ ablation in Table~3 to include the two extreme corners $\alpha{=}0.0$ (pure correctness) and $\alpha{=}1.0$ (pure entropy), addressing the concern that ``$\alpha{=}0.1$'' in the main table is not strictly correctness-only. Results in Table~\ref{tab:alpha_extended} confirm the trend: at $\alpha{=}0.0$ the difficulty signal is purely based on correctness, which under $G{=}8$ rollouts gives only coarse 12.5\% increments and the longest outputs (Avg.\ Len.\ 196, AES 0.76); at $\alpha{=}1.0$, removing the correctness anchor leads to the largest accuracy drop (Avg.\ Acc.\ 52.65\%) because the model can no longer distinguish lucky guesses from true mastery. The balanced $\alpha{=}0.5$ remains the best operating point (AES 0.93).
\begin{table*}[h]
\centering
\caption{Extended $\alpha$ ablation including the extreme corners $\alpha{=}0.0$ and $\alpha{=}1.0$. The balanced $\alpha{=}0.5$ achieves the best AES.}
\label{tab:alpha_extended}
\resizebox{\textwidth}{!}{%
\begin{tabular}{lcccccccccccccccc}
\toprule
\multirow{2}{*}{$\alpha$} & \multicolumn{2}{c}{MedQA} & \multicolumn{2}{c}{MedMCQA} & \multicolumn{2}{c}{PubMedQA} & \multicolumn{2}{c}{MMLU-ProM} & \multicolumn{2}{c}{GPQA-M} & \multicolumn{2}{c}{MedXpert} & \multicolumn{2}{c}{Avg} & \multirow{2}{*}{AES} \\
\cmidrule(lr){2-3} \cmidrule(lr){4-5} \cmidrule(lr){6-7} \cmidrule(lr){8-9} \cmidrule(lr){10-11} \cmidrule(lr){12-13} \cmidrule(lr){14-15}
 & Acc. & Len. & Acc. & Len. & Acc. & Len. & Acc. & Len. & Acc. & Len. & Acc. & Len. & Acc. & Len. & \\
\midrule
0.0 (correctness-only)  & 65.74 & 137 & 59.21 & 184 & 74.20 & 137 & 66.35 & 271 & 50.86 & 284 & 15.66 & 165 & 55.34 & 196 & 0.76 \\
0.1 (accuracy-focused)  & 66.06 & 121 & 59.02 & 153 & 74.30 & 116 & 66.19 & 195 & 50.51 & 258 & 15.39 & 142 & 55.25 & 164 & 0.82 \\
\rowcolor{highlightrow_yellow}
0.5 (balanced)          & 68.34 & 79  & 58.74 & 109 & 73.50 & 62  & 66.45 & 121 & 48.72 & 175 & 14.16 & 88  & 54.99 & 106 & \textbf{0.93} \\
0.9 (uncertainty-focused)& 63.47 & 44 & 58.12 & 216 & 73.60 & 10  & 63.19 & 50  & 46.92 & 69  & 14.08 & 30  & 53.23 & 70  & 0.90 \\
1.0 (entropy-only)      & 62.85 & 38  & 57.34 & 224 & 73.20 & 8   & 62.41 & 41  & 46.26 & 57  & 13.81 & 24  & 52.65 & 65  & 0.88 \\
\bottomrule
\end{tabular}}
\end{table*}

\subsection{Sensitivity Analysis of TOP-$K$}
To assess the impact of the hyperparameter TOP-$K$, we conducted ablation experiments using the Qwen2.5-Instruct-7B model. The results shown in Table~\ref{tab:ablation_k} indicate that variations in the TOP-$K$ value lead to stable efficiency improvements in AdaThink-Med. However, both very large and very small values of TOP-$K$ result in suboptimal efficiency gains. This occurs because smaller $K$ values reduce the number of sampled tokens, leading to inaccurate difficulty estimations, while larger $K$ values introduce an averaging effect that dilutes the model's entropy due to the presence of plain tokens.

\begin{table*}[ht]
\centering
{
\caption{Ablation results on the TOP-$K$ hyperparameter for entropy calculation.}
\label{tab:ablation_k}
\resizebox{\textwidth}{!}{%
\begin{tabular}{l|cc|cc|cc|cc|cc|cc|cc|c}
\toprule
\multirow{2}{*}{{$K$}} &\multicolumn{2}{c|}{{MedQA}}&\multicolumn{2}{c|}{{MedMCQA}} &\multicolumn{2}{c|}{{PubMedQA}} & \multicolumn{2}{c|}{{MMLU-ProM}} & \multicolumn{2}{c|}{{GPQA-M}} & \multicolumn{2}{c|}{{MedXpert}} & \multicolumn{2}{c|}{{Avg.}} &\multirow{2}{*}{{AES}}\\
\cline{2-15}
 & {Acc.} & {Len.} & {Acc.} & {Len.} & {Acc.} & {Len.} & {Acc.} & {Len.} & {Acc.} & {Len.}& {Acc.} & {Len.} & {Acc.} & {Len.} & \\
\midrule
- & 63.86 & 490 & 65.68 & 359 & 73.10 & 386 & 62.23 & 523 & 46.41 & 562 & 12.28 & 585 & 52.44 & 484 & - \\
5 & 67.83 & 102 & 67.53 & 125 & 74.22 & 87  & 65.93 & 131 & 46.96 & 221 & 13.95 & 102 & 54.41 & 128 & +0.85 \\
20 & 68.34 & 79  & 68.74 & 109 & 73.50 & 62  & 66.40 & 121 & 48.72 & 175 & 14.16 & 109 & 54.99 & 106 & +0.93 \\
50 & 67.55 & 141 & 68.35 & 127 & 73.92 & 105 & 68.32 & 143 & 46.79 & 190 & 14.24 & 117 & 54.86 & 137 & +0.86 \\
\hline
\end{tabular}}
}
\end{table*}

\subsection{Sensitivity Analysis of the AES Metric}
\label{appendix:aes_sensitivity}

To ensure a fair and consistent comparison with prior work in adaptive reasoning, we adopted the Accuracy-Efficiency Score (AES) using the standard hyperparameters ($\alpha=1, \beta=3, \gamma=5$) as established in recent studies~\citep{yi2025shorterbetter, luo2025o1}. This metric was designed to provide a comprehensive assessment of the trade-off between reducing computational costs and maintaining model performance.

However, to address potential concerns regarding the sensitivity of the AES metric to specific hyperparameter choices, we conducted a robustness analysis. We evaluated AdaThink-Med and baseline models under a diverse set of AES configurations, systematically varying the weights for length reduction ($\alpha$), accuracy gain ($\beta$), and accuracy loss penalty ($\gamma$).

The results, presented in Table~\ref{tab:aes_sensitivity}, demonstrate that AdaThink-Med consistently achieves the highest AES across all tested hyperparameter combinations. Whether the metric configuration prioritizes strict accuracy preservation (high $\gamma$) or aggressive efficiency gains (high $\alpha$), our method maintains its superiority. These findings confirm that the reported performance advantage is intrinsic to the adaptive reasoning mechanism of AdaThink-Med and is robust to variations in the evaluation metric.

\begin{table*}[t]
\centering
\caption{Sensitivity analysis of the Accuracy-Efficiency Score (AES) under varying hyperparameter settings. $\Delta L$ and $\Delta A$ represent the relative reduction in length and change in accuracy, respectively. The standard configuration used in the main text is $\alpha=1, \beta=3, \gamma=5$. AdaThink-Med consistently outperforms baselines across all weighting schemes. Best results are in \textbf{bold}.}
\label{tab:aes_sensitivity}
\resizebox{\textwidth}{!}{%
\begin{tabular}{lcccccccccc}
\toprule
\multirow{2}{*}{Model} & \multirow{2}{*}{Avg Acc.} & \multirow{2}{*}{Avg Len.} & \multirow{2}{*}{$\Delta L$} & \multirow{2}{*}{$\Delta A$} & \multicolumn{6}{c}{AES Configurations ($\alpha, \beta, \gamma$)} \\
\cmidrule(lr){6-11}
 & & & & & 1, 3, 5 & 1, 5, 5 & 1, 2, 3 & 1, 2, 10 & 0.5, 2, 1 & 0.1, 1, 2 \\
\midrule
\multicolumn{11}{c}{\textit{Llama-3.1-8B Backbone}} \\
\midrule
Llama-3.1-8B-Instruct [CoT] & 54.25 & 425.27 & - & - & - & - & - & - & - & - \\
Llama-3.1-8B-Instruct & 48.93 & 254.38 & 0.40 & -0.10 & -0.09 & -0.09 & 0.11 & -0.58 & 0.10 & -0.16 \\
Llama-3.1-8B-Instruct [GRPO] & 56.72 & 409.88 & 0.04 & 0.05 & 0.17 & 0.26 & 0.13 & 0.17 & 0.11 & 0.05 \\
Kimi & 55.65 & 124.89 & 0.71 & 0.03 & 0.78 & 0.84 & 0.76 & 0.78 & 0.40 & 0.10 \\
CosFn & 38.10 & 973.64 & -1.29 & -0.30 & -2.78 & -2.78 & -2.18 & -4.27 & -0.94 & -0.72 \\
DAST & 52.71 & 51.37 & 0.88 & -0.03 & 0.74 & 0.74 & 0.79 & 0.60 & 0.41 & 0.03 \\
ShortBetter & 55.94 & 167.38 & 0.61 & 0.03 & 0.70 & 0.76 & 0.67 & 0.70 & 0.37 & 0.09 \\
\textbf{AdaThink-Med-Llama} & 55.59 & 63.81 & 0.85 & 0.02 & \textbf{0.92} & \textbf{0.97} & \textbf{0.90} & \textbf{0.92} & \textbf{0.47} & \textbf{0.11} \\
\midrule
\multicolumn{11}{c}{\textit{Qwen2.5-7B Backbone}} \\
\midrule
Qwen-Instruct-7B [CoT] & 52.44 & 484.08 & - & - & - & - & - & - & - & - \\
Qwen-Instruct-7B & 48.04 & 278.24 & 0.43 & -0.08 & 0.01 & 0.01 & 0.17 & -0.41 & 0.13 & -0.13 \\
Qwen-Instruct-7B [GRPO] & 54.74 & 496.62 & -0.03 & 0.04 & 0.11 & 0.19 & 0.06 & 0.11 & 0.07 & 0.04 \\
Kimi & 55.41 & 298.09 & 0.38 & 0.06 & 0.55 & 0.67 & 0.50 & 0.55 & 0.31 & 0.10 \\
CosFn & 38.10 & 973.64 & -1.01 & -0.27 & -2.38 & -2.38 & -1.83 & -3.75 & -0.78 & -0.65 \\
DAST & 54.54 & 283.01 & 0.42 & 0.04 & 0.54 & 0.62 & 0.50 & 0.54 & 0.29 & 0.08 \\
ShortBetter & 54.28 & 335.19 & 0.31 & 0.04 & 0.41 & 0.48 & 0.38 & 0.41 & 0.22 & 0.07 \\
\textbf{AdaThink-Med-Qwen} & 54.99 & 105.71 & 0.78 & 0.05 & \textbf{0.93} & \textbf{1.02} & \textbf{0.88} & \textbf{0.93} & \textbf{0.49} & \textbf{0.13} \\
\bottomrule
\end{tabular}%
}
\end{table*}

\begin{table*}[h]
\centering
\caption{Experimental results on using adaptive thinking methods on non-R1 reasoning models, here we use HuatuoGPT-o1, which first SFT with CoT data and then RL with PPO.}
\label{tab:ppo_reasoning_model}
\resizebox{\textwidth}{!}{%
\begin{tabular}{lccccccccccccccccr}
\toprule
\multirow{2}{*}{Model} & \multicolumn{2}{c}{MedQA} & \multicolumn{2}{c}{MedMCQA} & \multicolumn{2}{c}{PubMedQA} & \multicolumn{2}{c}{MMLU-ProM} & \multicolumn{2}{c}{GPQA-M} & \multicolumn{2}{c}{MedXpert} & \multicolumn{2}{c}{Avg} & \multirow{2}{*}{AES} \\
 & Acc. & Len. & Acc. & Len. & Acc. & Len. & Acc. & Len. & Acc. & Len. & Acc. & Len. & Acc. & Len. & \\
\midrule
HuatuoGPT-o1-8B & 76.35 & 568 & 62.82 & 445 & 79.80 & 445 & 63.71 & 521 & 54.35 & 585 & 17.06 & 601 & 59.02 & 527 & -- \\
Kimi1.5~\citep{team2025kimi} & 71.01 & 373 & 62.87 & 302 & 77.40 & 288 & 67.04 & 340 & 50.26 & 401 & 14.49 & 363 & 57.18 & 345 & 0.19 \\
CosFn~\citep{yeo2025demystifying} & 68.42 & 260 & 62.63 & 126 & 77.30 & 159 & 65.60 & 220 & 48.97 & 187 & 14.24 & 246 & 56.20 & 200 & 0.38 \\
DAST~\citep{shen2025dast} & 71.17 & 730 & 61.01 & 619 & 76.70 & 578 & 64.36 & 697 & 45.90 & 794 & 15.02 & 765 & 55.69 & 697 & -0.60 \\
ShortBetter~\citep{yi2025shorterbetter} & 71.64 & 336 & 63.18 & 271 & 78.40 & 258 & 66.12 & 299 & 49.74 & 353 & 14.45 & 321 & 57.26 & 306 & 0.27 \\
\textbf{AdaThink-Med-HuatuoGPT-o1} & 72.74 & 183 & 63.54 & 192 & 76.60 & 72 & 66.96 & 204 & 50.77 & 344 & 13.88 & 221 & 57.41 & 203 & 0.48 \\
\bottomrule
\end{tabular}
}
\end{table*}

\subsection{Effectiveness of AdaThink-Med on Non-R1-Like Models}

The adaptive thinking experiments in the main paper are performed on R1-like models, where reasoning capability is acquired through RL training with GRPO. To further validate our approach, we conduct additional experiments on other reasoning models. Our implementation follows the official setup of HuatuoGPT-o1~\citep{chen2024huatuogpt-o1}, which employs a large language model (LLM) as a judgment mechanism. Specifically, we use HuatuoGPT-o1-7B~\citep{chen2024huatuogpt-o1} as our base model, which is built upon the Qwen2.5-Instruct-7B backbone and trained in two stages: first through supervised fine-tuning (SFT) on 20,000 long-form CoT samples, followed by reinforcement learning via proximal policy optimization (PPO) on 20,000 verifiable medical question-answering samples. All 40,000 samples are converted into open-ended QA format and sourced from the training splits of MedQA-USMLE and MedMCQA. The 20,000 PPO samples are further utilized in our subsequent adaptive GRPO training. For verification, we adopt the distilled verifier released by HuatuoGPT-o1, which is trained using 20,000 scoring samples distilled from GPT-4o.

We compare several RL-based length calibration methods, and the results are summarized in Table~\ref{tab:ppo_reasoning_model}. Our findings demonstrate that AdaThink-Med achieves the highest efficiency improvement. Specifically, AdaThink-Med reduces the average output length by a factor of 2.6, from 527 to 203, while incurring only an average performance degradation of 1.59\%. In comparison, the second-best method, CosFn, compresses the average output length from 527 to 200, but with a larger performance degradation of 2.78\%. Overall, these results confirm the effectiveness of AdaThink-Med in enhancing the inference efficiency of diverse reasoning models.

\begin{figure*}
    \centering
    \includegraphics[width=0.8\linewidth]{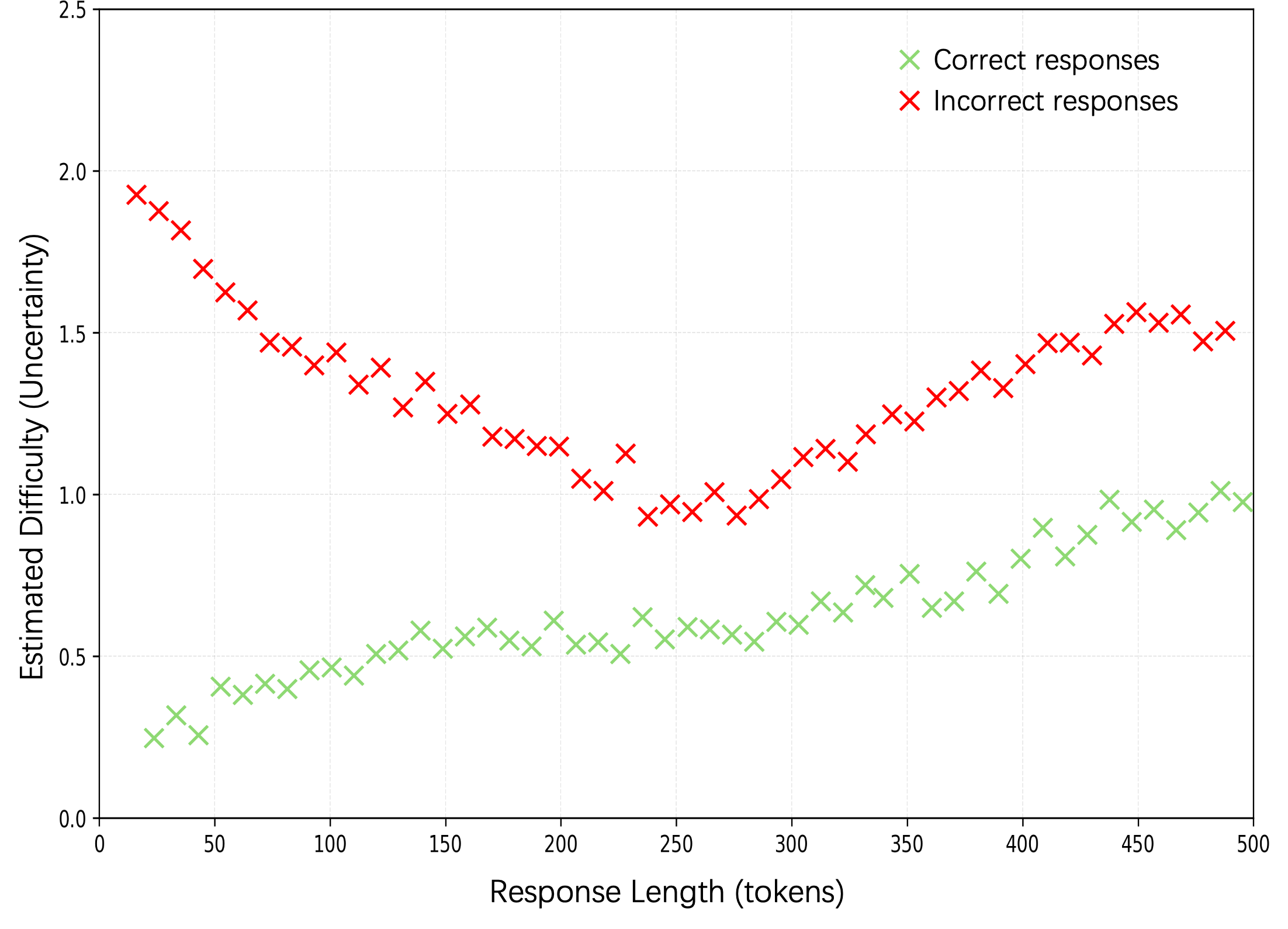}
    \caption{{Estimated difficulty vs. response length.}}
    \label{fig:infer_entropy}
\end{figure*}

\subsection{Estimated Difficulty vs.\ Response Length Analysis}
Additionally, we have included the requested plot of response length vs. estimated difficulty in Fig.~\ref{fig:infer_entropy}. The overall trend observed is as follows: for correct test samples, as the response length increases, the uncertainty also rises, indicating an increase in difficulty. For incorrect samples, both short and long response lengths show high difficulty, with short responses exhibiting a more pronounced rise in uncertainty. This suggests that more reasoning or cognitive effort is required in such cases. Notably, in the region of medium response lengths, both uncertainty measures level off, indicating that these problems are challenging to optimize using a single mode.

\subsection{Generalization to Large-Scale Reasoning Models}
\label{appendix:generalization_qwen3}

To assess the generalizability of AdaThink-Med to larger-scale and state-of-the-art architectures, we extended our experimental evaluation to include Qwen-3-32B. As a reasoning-intensive model optimized through reinforcement learning, Qwen-3-32B serves as a robust benchmark for evaluating adaptive thinking capabilities in advanced LLMs. We directly applied the AdaThink-Med framework to this model without architectural modifications to test its adaptability.

The empirical results are summarized in Table~\ref{tab:qwen3_results}. We observe that AdaThink-Med consistently achieves substantial reductions in inference length compared to the baseline and other length-control methods. Specifically, the average output length is reduced from 1888 tokens to 954 tokens, representing a compression of approximately 49\%. Importantly, this efficiency gain comes with minimal performance degradation, resulting in the highest Accuracy-Efficiency Score (AES) of 0.45. These findings further corroborate the scalability and effectiveness of AdaThink-Med when applied to newer, larger parameter models that possess inherent reasoning capabilities.

\begin{table*}[h]
\centering
\caption{Performance evaluation on the large-scale Qwen-3-32B model. AdaThink-Med achieves superior efficiency gains while maintaining competitive accuracy across six medical benchmarks. The best AES scores are highlighted in \textbf{bold}.}
\label{tab:qwen3_results}
\resizebox{\textwidth}{!}{%
\begin{tabular}{lccccccccccccccc}
\toprule
\multirow{2}{*}{Model} & \multicolumn{2}{c}{MedQA} & \multicolumn{2}{c}{MedMCQA} & \multicolumn{2}{c}{PubMedQA} & \multicolumn{2}{c}{MMLU-ProM} & \multicolumn{2}{c}{GPQA-M} & \multicolumn{2}{c}{MedXpert} & \multicolumn{2}{c}{Avg} & \multirow{2}{*}{AES} \\
\cmidrule(lr){2-3} \cmidrule(lr){4-5} \cmidrule(lr){6-7} \cmidrule(lr){8-9} \cmidrule(lr){10-11} \cmidrule(lr){12-13} \cmidrule(lr){14-15}
 & Acc. & Len. & Acc. & Len. & Acc. & Len. & Acc. & Len. & Acc. & Len. & Acc. & Len. & Acc. & Len. & \\
\midrule
Qwen3-32B & 87.50 & 1679 & 72.34 & 1473 & 73.50 & 722 & 79.67 & 1770 & 68.46 & 3009 & 26.41 & 2676 & 67.98 & 1888 & 0.00 \\
Kimi & 84.23 & 1099 & 69.93 & 1002 & 73.00 & 656 & 79.45 & 1335 & 67.25 & 1776 & 25.37 & 2108 & 66.54 & 1329 & +0.19 \\
CosFn & 85.60 & 1523 & 72.12 & 1355 & 72.20 & 585 & 78.53 & 1557 & 66.25 & 2553 & 24.45 & 2330 & 66.53 & 1651 & +0.02 \\
DAST & 86.53 & 1345 & 71.88 & 1154 & 73.10 & 724 & 76.53 & 1553 & 64.32 & 1889 & 24.32 & 2235 & 66.11 & 1513 & +0.06 \\
ShortBetter & 84.30 & 1427 & 70.15 & 1238 & 72.50 & 669 & 76.78 & 1476 & 67.24 & 1920 & 24.99 & 2140 & 65.99 & 1478 & +0.07 \\
AdaThink-Med & 86.20 & 1120 & 72.55 & 893 & 75.40 & 256 & 77.76 & 785 & 66.24 & 1431 & 25.88 & 1237 & 67.34 & 954 & \textbf{+0.45} \\
\bottomrule
\end{tabular}%
}
\end{table*}

\section{Further Analyses}

\subsection{Real-World Open-Ended Clinical Evaluation: Additional Details}
\label{app:clinical_open_ended}
The open-ended evaluation on 796 anonymized real-world cardiac stroke cases is summarized in the main text (Section~\ref{sec:results}, Table~\ref{tab:clinical_open_ended}). The dataset was curated in collaboration with four hospitals and de-identified prior to use. Each model received the raw clinical record (history of present illness, past medical history, examination, and relevant imaging notes) and was prompted to produce a free-text \emph{Diagnosis} and a free-text \emph{Treatment Plan}. Three board-certified clinicians blindly scored every generation along three axes: \emph{Accuracy} (factual correctness against the gold standard), \emph{Redundancy} (proportion of clinically non-essential content; lower is better), and \emph{Clinical ICA} (logical soundness and completeness of the reasoning). Disagreements were resolved by consensus discussion. As reported in the main text, AdaThink-Med achieves the highest Accuracy and Clinical ICA on both subtasks while keeping Redundancy at 18.6\% (Diagnosis) and 24.3\% (Treatment). Greedy length-reduction baselines (Kimi, ShortBetter) further reduce Redundancy but at the cost of Clinical ICA dropping below 70\%, indicating systematic over-pruning of clinically essential reasoning steps---a failure mode our uncertainty-guided mechanism explicitly prevents by tying compression to model confidence.

\subsection{Repetition Penalty Ablation on Llama}
\label{app:rep_pen_ablation}
As briefly noted in Appendix~\ref{sec_supp_imple_details}, training Llama-3.1-8B with adaptive length calibration requires an additional $n$-gram repetition penalty to prevent reward hacking via degenerate token repetition. Table~\ref{tab:rep_pen_llama} quantifies this: removing the repetition penalty causes \emph{complete} training collapse rather than mild degradation. Without the penalty the policy enters degenerate $n$-gram loops, padding outputs to the maximum token budget without producing valid reasoning---consistent with prior observations on Llama backbones~\citep{yeo2025demystifying}. Qwen-2.5 backbones inherently resist this failure mode and do \emph{not} require the penalty, so we apply it only on Llama for fairness across backbones.
\begin{table}[h]
\centering
\caption{Effect of removing the $n$-gram repetition penalty on Llama-3.1-8B during adaptive RL. Average accuracy and length over the six medical benchmarks. Without the penalty, training collapses into degenerate repetition.}
\label{tab:rep_pen_llama}
\begin{tabular}{lccc}
\toprule
\textbf{Configuration} & \textbf{Avg.\ Acc.\ (\%)} & \textbf{Avg.\ Len.} & \textbf{AES} \\
\midrule
Standard GRPO-Llama                          & 56.72 & 410  & $-0.09$ \\
AdaThink-Llama without repetition penalty    & 13.26 & 1024 & $-3.08$ \\
AdaThink-Llama with repetition penalty       & 55.59 & 64   & $+0.92$ \\
\bottomrule
\end{tabular}
\end{table}

\subsection{Controlled Compute-Matched GRPO Baseline}
\label{app:grpo500}
Our staged training pipeline first runs vanilla GRPO for 300 steps, then applies adaptive length calibration for an additional 200 steps (500 total). To rule out the possibility that AdaThink-Med's improvements over standard GRPO simply reflect more total compute, we run a compute-matched control: extend vanilla GRPO from 300 to 500 steps without any length calibration. Results in Table~\ref{tab:grpo500} show that GRPO(500) yields negligible additional improvement over GRPO(300) in both accuracy and length, confirming that the observed gains stem from the adaptive length reward design rather than from longer training.
\begin{table}[h]
\centering
\caption{Compute-matched control: extending vanilla GRPO to the same total step count (500) as AdaThink-Med yields negligible change, confirming gains come from the reward design.}
\label{tab:grpo500}
\begin{tabular}{lcc}
\toprule
\textbf{Backbone} & \textbf{GRPO(300) Acc/Len} & \textbf{GRPO(500) Acc/Len} \\
\midrule
Llama-3.1-8B  & 56.72\% / 410 & 56.73\% / 408 \\
Qwen-2.5-7B   & 54.74\% / 497 & 54.71\% / 494 \\
\midrule
\textbf{AdaThink-Med (Llama)} & \multicolumn{2}{c}{55.59\% / 64} \\
\textbf{AdaThink-Med (Qwen)}  & \multicolumn{2}{c}{54.99\% / 106} \\
\bottomrule
\end{tabular}
\end{table}

\begin{table*}[ht]
\centering
{
\caption{Experimental results showing the robustness of AdaThink-Med under low-quality output conditions.}
\label{tab: robust}
\begin{tabular}{l|cc|cc|cc|cc|cc}
\toprule
\multirow{2}{*}{{Method}} & \multicolumn{2}{c|}{{PubMedQA}} & \multicolumn{2}{c|}{{MMLU-ProM}} & \multicolumn{2}{c|}{{GPQA-M}} & \multicolumn{2}{c|}{{MedXpert}} & \multicolumn{2}{c}{{Avg.}} \\
\cline{2-11}
 & {Acc.} & {Len.} & {Acc.} & {Len.} & {Acc.} & {Len.} & {Acc.} & {Len.} & {Acc.} & {Len.} \\
\midrule
Qwen-GRPO & 72.30 & 396 & 66.12 & 535 & 48.46 & 575 & 14.48 & 592 & 50.34 &524 \\
Kimi & 71.20 & 230 & 65.23 & 328 & 46.52 & 376 & 14.31 & 366 & 49.34 &325 \\
CosFn & 70.90 & 301 & 64.17 & 428 & 47.66 & 351 & 13.87 & 353 & 49.15 &358 \\
DAST & 72.30 & 298 & 64.88 & 535 & 47.35 & 507 & 13.26 & 498 & 49.45 &460 \\
ShortBetter & 70.50 & 354 & 63.15 & 297 & 45.23 & 330 & 13.26 & 320 & 48.04 &325 \\
AdaThink-Med & 72.10 & 89 & 65.38 & 115 & 47.51 & 212 & 14.33 & 108 & 49.83 &131 \\
\bottomrule
\end{tabular}}
\end{table*}

\subsection{Robustness to Low-Quality Outputs}

To assess the robustness of our algorithm to low-quality outputs, we conducted experiments using a low-quality dataset. Specifically, after training the Qwen2.5-Instruct-7B model with GRPO, we performed multiple rounds of response reasoning for the same question in the training set. We then ranked the samples based on accuracy and selected both the fully incorrect samples and some low-accuracy samples to form our low-quality dataset. These samples, which generate high uncertainty, present a significant challenge for training. As shown in Table~\ref{tab: robust}, the experimental results indicate that AdaThink-Med consistently outperforms other models under these challenging conditions, demonstrating the robustness of our algorithm.

\subsection{VQA Generalization Ability}
To evaluate the effectiveness of AdaThink-Med in multi-modal scenarios, we conducted a clinical free-text generation task. Specifically, we included experimental results from the radiology (X-ray) report generation task, which is a Visual Question Answering (VQA) task. The MIMIC-CXR-2.0~\citep{johnson2019mimic} dataset was used for validation. In the test set, we employed LLM-as-judge to assess the generated reports and compare them to the ground truth in two key aspects: the completeness of the report and its conciseness. As shown in Table~\ref{tab:vqa_res}, the results demonstrate that AdaThink-Med effectively reduces the report length while maintaining completeness, highlighting its exceptional performance.

\subsection{Sensitivity Analysis of Entropy Estimation via Mixed LLMs}
\label{appendix:entropy_sensitivity}

To evaluate the robustness of our entropy-based difficulty estimator and explore potential performance gains from model ensembling, we conducted a sensitivity analysis by utilizing a mixture of LLMs for entropy estimation. In our main experiments, the entropy used for difficulty estimation is derived from the target model itself (self-estimation). In this supplementary experiment, we introduce a cross-model estimation strategy.

Specifically, we define \textbf{AdaThink-Med-Llama(+Qwen)} as the Llama-3.1-8B-Instruct model where the adaptive reasoning process is guided by entropy signals derived from a GRPO-enhanced Qwen2.5-7B-Instruct model. Conversely, \textbf{AdaThink-Med-Qwen(+Llama)} utilizes the Llama model to estimate uncertainty for the Qwen backbone. This approach leverages the diversity of probability distributions across different LLM architectures to provide a more comprehensive and stable entropy estimate.

The experimental results are presented in Table~\ref{tab:mixed_llm_entropy}. We observe that employing mixed LLMs for entropy estimation yields consistent improvements over the single-model baseline:

\begin{enumerate}
    \item \textbf{Improved Accuracy and Efficiency:} The mixed-model approach results in more accurate difficulty estimation, allowing the model to prune redundant reasoning steps more aggressively without sacrificing correctness. For instance, AdaThink-Med-Llama(+Qwen) achieves an Accuracy-Efficiency Score (AES) of \textbf{1.06}, surpassing the single-model version (0.92).
    \item \textbf{Mitigation of Performance Drops:} This strategy effectively addresses the performance fluctuations observed in specific datasets. Notably, on the MedQA dataset, where the single-model Llama variant experienced a slight accuracy trade-off, the Qwen-guided variant recovers performance significantly (from 67.00\% to 74.21\%) while further reducing the average output length (from 36 to 33 tokens).
\end{enumerate}

These findings suggest that our uncertainty-guided length calibration mechanism is not only robust but can be further enhanced through ensemble-based entropy estimation, paving the way for more efficient and accurate medical reasoning systems.

\begin{table*}[h]
\centering
\caption{Performance comparison of AdaThink-Med using single vs. mixed LLMs for entropy estimation across six medical benchmarks. ``+Model'' indicates the source of external entropy guidance. The best AES scores are highlighted in \textbf{bold}.}
\label{tab:mixed_llm_entropy}
\resizebox{\textwidth}{!}{%
\begin{tabular}{lccccccccccccccc}
\toprule
\multirow{2}{*}{Model} & \multicolumn{2}{c}{MedQA} & \multicolumn{2}{c}{MedMCQA} & \multicolumn{2}{c}{PubMedQA} & \multicolumn{2}{c}{MMLU-ProM} & \multicolumn{2}{c}{GPQA-M} & \multicolumn{2}{c}{MedXpert} & \multicolumn{2}{c}{Avg} & \multirow{2}{*}{AES} \\
\cmidrule(lr){2-3} \cmidrule(lr){4-5} \cmidrule(lr){6-7} \cmidrule(lr){8-9} \cmidrule(lr){10-11} \cmidrule(lr){12-13} \cmidrule(lr){14-15}
 & Acc. & Len. & Acc. & Len. & Acc. & Len. & Acc. & Len. & Acc. & Len. & Acc. & Len. & Acc. & Len. & \\
\midrule
\multicolumn{16}{c}{\textit{Llama-3.1-8B Backbone}} \\
\midrule
Llama-3.1-8B-Instruct [CoT] & 68.18 & 461 & 57.32 & 309 & 78.20 & 296 & 60.06 & 457 & 45.64 & 485 & 16.08 & 545 & 54.25 & 425 & - \\
Llama-3.1-8B-Instruct & 54.75 & 302 & 54.69 & 169 & 77.00 & 113 & 57.00 & 256 & 35.64 & 345 & 14.48 & 342 & 48.93 & 254 & -0.09 \\
Llama-3.1-8B-Instruct [GRPO] & 72.50 & 438 & 61.60 & 299 & 78.50 & 290 & 63.51 & 442 & 46.92 & 461 & 17.26 & 529 & 56.72 & 410 & 0.17 \\
Kimi & 70.62 & 145 & 60.93 & 100 & 77.50 & 92 & 63.71 & 164 & 45.38 & 118 & 15.75 & 131 & 55.65 & 125 & 0.78 \\
CosFn & 37.94 & 1007 & 49.29 & 919 & 75.10 & 885 & 23.32 & 998 & 32.82 & 1012 & 10.12 & 1020 & 38.10 & 974 & -2.78 \\
DAST & 60.64 & 30 & 56.90 & 34 & 76.20 & 39 & 59.48 & 71 & 47.69 & 53 & 15.35 & 82 & 52.71 & 51 & 0.74 \\
ShortBetter & 71.48 & 175 & 60.62 & 134 & 78.60 & 118 & 62.99 & 188 & 45.89 & 219 & 16.08 & 171 & 55.94 & 167 & 0.70 \\
AdaThink-Med-Llama & 67.00 & 36 & 60.41 & 73 & 78.40 & 54 & 63.58 & 85 & 47.94 & 100 & 16.20 & 35 & 55.59 & 64 & 0.92 \\
\textbf{AdaThink-Med-Llama(+Qwen)} & \textbf{74.21} & \textbf{33} & \textbf{62.35} & \textbf{60} & 78.22 & \textbf{32} & \textbf{63.59} & \textbf{78} & \textbf{48.21} & \textbf{88} & \textbf{18.77} & \textbf{30} & \textbf{57.56} & \textbf{54} & \textbf{1.06} \\
\midrule
\multicolumn{16}{c}{\textit{Qwen2.5-7B Backbone}} \\
\midrule
Qwen-Instruct-7B [CoT] & 63.86 & 490 & 56.68 & 359 & 73.10 & 386 & 62.28 & 523 & 46.41 & 562 & 12.28 & 585 & 52.44 & 484 & - \\
Qwen-Instruct-7B & 54.28 & 313 & 53.43 & 196 & 72.70 & 129 & 56.67 & 292 & 38.71 & 448 & 12.44 & 291 & 48.04 & 278 & 0.01 \\
Qwen-Instruct-7B [GRPO] & 67.55 & 504 & 59.52 & 378 & 72.30 & 396 & 66.12 & 535 & 48.46 & 575 & 14.48 & 592 & 54.74 & 497 & 0.11 \\
Kimi & 69.99 & 285 & 58.95 & 225 & 75.00 & 236 & 66.71 & 338 & 46.92 & 376 & 14.89 & 330 & 55.41 & 298 & 0.55 \\
CosFn & 37.94 & 1007 & 49.29 & 919 & 75.10 & 885 & 23.32 & 998 & 32.82 & 1012 & 10.12 & 1020 & 38.10 & 974 & -2.38 \\
DAST & 65.90 & 260 & 59.67 & 327 & 69.70 & 153 & 65.86 & 205 & 50.51 & 466 & 15.59 & 288 & 54.54 & 283 & 0.54 \\
ShortBetter & 67.94 & 333 & 58.90 & 250 & 71.70 & 260 & 65.86 & 376 & 46.92 & 405 & 14.36 & 387 & 54.28 & 335 & 0.41 \\
AdaThink-Med-Qwen & 68.34 & 79 & 58.74 & 109 & 73.50 & 62 & 66.45 & 121 & 48.72 & 175 & 14.16 & 88 & 54.99 & 106 & 0.93 \\
\textbf{AdaThink-Med-Qwen(+Llama)} & \textbf{69.12} & \textbf{61} & \textbf{59.87} & \textbf{95} & 73.50 & \textbf{35} & \textbf{66.88} & \textbf{89} & 48.58 & \textbf{143} & 14.33 & \textbf{75} & \textbf{55.38} & \textbf{83} & \textbf{1.00} \\
\bottomrule
\end{tabular}%
}
\end{table*}

\begin{table*}[ht]
\centering
{
\caption{Experimental results for the radiology report generation task. Completeness is measured by the extent to which the generated report covers all required aspects, and redundancy reflects the level of unnecessary components in the generated text.}
\label{tab:vqa_res}
\begin{tabular}{l|c|c}
\hline
\textbf{Method} & \textbf{Completeness ↑} & \textbf{Redundancy ↓} \\
\hline
Kimi & 66.2 & 42.5 \\
CosFn & 34.5 & 54.1 \\
DAST & 46.7 & 50.7 \\
ShortBetter & 55.3 & 45.2 \\
AdaThink-Med & \textbf{78.4} & \textbf{22.5} \\
\hline
\end{tabular}
}
\end{table*}

\subsection{Scalability Analysis on Long-Chain Reasoning Models}
\label{appendix:long_chain_scalability}

A critical consideration in efficient reasoning is the domain-specific nature of token consumption. In mathematical reasoning benchmarks (e.g., MATH500), high token counts are often driven by the verbose tokenization of numerical symbols and formulas. In contrast, medical reasoning relies primarily on plain text for differential diagnosis and evidence-based analysis, which is naturally more token-efficient. However, complex clinical queries still demand extended reasoning chains comparable to those found in mathematical tasks. Consequently, an effective compression strategy must be adaptive, capable of handling the high variance between simple factoid questions and intricate diagnostic scenarios.

To validate the scalability of AdaThink-Med on models that inherently generate extended reasoning chains, we conducted additional evaluations using HuatuoGPT-o1-8B and m1-7B. Unlike the standard instruction-tuned models used in the main experiments, these models are specifically optimized for long-context reasoning and exhibit significantly higher average output lengths.

The results, presented in Table~\ref{tab:long_chain_models}, demonstrate the effectiveness of our approach in this high-token regime. AdaThink-Med successfully compresses the reasoning paths of the m1-7B model from an average of 2451 tokens to 1100 tokens, achieving a reduction of approximately 55\% while maintaining an average accuracy of 58.03\%. Similarly, for HuatuoGPT-o1-8B, our method reduces the average length from 527 to 203 tokens with minimal impact on performance. The consistently higher Accuracy-Efficiency Scores (AES) compared to baselines such as Kimi, CosFn, and ShortBetter confirm that our uncertainty-guided length calibration mechanism is robust and scalable, effectively optimizing models with diverse intrinsic reasoning length distributions.

\begin{table*}[h]
\centering
\caption{Performance comparison on long-chain reasoning models (HuatuoGPT-o1-8B and m1-7B). AdaThink-Med achieves significant length compression while maintaining competitive accuracy across six medical benchmarks. The best AES scores are highlighted in \textbf{bold}.}
\label{tab:long_chain_models}
\resizebox{\textwidth}{!}{%
\begin{tabular}{lccccccccccccccc}
\toprule
\multirow{2}{*}{Model} & \multicolumn{2}{c}{MedQA} & \multicolumn{2}{c}{MedMCQA} & \multicolumn{2}{c}{PubMedQA} & \multicolumn{2}{c}{MMLU-ProM} & \multicolumn{2}{c}{GPQA-M} & \multicolumn{2}{c}{MedXpert} & \multicolumn{2}{c}{Avg} & \multirow{2}{*}{AES} \\
\cmidrule(lr){2-3} \cmidrule(lr){4-5} \cmidrule(lr){6-7} \cmidrule(lr){8-9} \cmidrule(lr){10-11} \cmidrule(lr){12-13} \cmidrule(lr){14-15}
 & Acc. & Len. & Acc. & Len. & Acc. & Len. & Acc. & Len. & Acc. & Len. & Acc. & Len. & Acc. & Len. & \\
\midrule
\multicolumn{16}{c}{\textit{HuatuoGPT-o1-8B Backbone}} \\
\midrule
HuatuoGPT-o1-8B [CoT] & 76.35 & 568 & 62.82 & 445 & 79.80 & 445 & 63.71 & 521 & 54.35 & 585 & 17.06 & 601 & 59.02 & 527 & - \\
Kimi & 71.01 & 373 & 62.87 & 302 & 77.40 & 288 & 67.04 & 340 & 50.26 & 401 & 14.49 & 363 & 57.18 & 345 & 0.19 \\
CosFn & 68.42 & 260 & 62.63 & 126 & 77.30 & 159 & 65.60 & 220 & 48.97 & 187 & 14.24 & 246 & 56.20 & 200 & 0.38 \\
DAST & 71.17 & 730 & 61.01 & 619 & 76.70 & 578 & 64.36 & 697 & 45.90 & 794 & 15.02 & 765 & 55.69 & 697 & -0.60 \\
ShortBetter & 71.64 & 336 & 63.18 & 271 & 78.40 & 258 & 66.12 & 299 & 49.74 & 353 & 14.45 & 321 & 57.26 & 306 & 0.27 \\
AdaThink-Med & 72.74 & 183 & 63.54 & 192 & 76.60 & 72 & 66.91 & 204 & 50.77 & 344 & 13.88 & 221 & 57.41 & 203 & \textbf{0.48} \\
\midrule
\multicolumn{16}{c}{\textit{m1-7B Backbone}} \\
\midrule
m1-7B [CoT] & 75.01 & 2161 & 62.32 & 1749 & 74.20 & 1080 & 68.07 & 2564 & 51.53 & 3887 & 18.28 & 3265 & 58.24 & 2451 & - \\
Kimi & 71.23 & 1425 & 61.26 & 992 & 72.50 & 856 & 65.56 & 1335 & 50.58 & 2069 & 15.53 & 1974 & 56.11 & 1442 & 0.23 \\
CosFn & 72.58 & 1566 & 60.57 & 1011 & 72.70 & 774 & 67.99 & 1218 & 51.35 & 1655 & 16.23 & 1563 & 56.90 & 1298 & 0.36 \\
DAST & 70.87 & 2445 & 58.43 & 1995 & 70.10 & 1243 & 65.33 & 2655 & 48.33 & 4096 & 14.43 & 4096 & 54.58 & 2755 & -0.44 \\
ShortBetter & 73.55 & 1359 & 60.58 & 927 & 71.40 & 839 & 65.37 & 1290 & 50.12 & 1847 & 14.99 & 1885 & 56.00 & 1358 & 0.25 \\
AdaThink-Med & 74.15 & 982 & 63.35 & 876 & 73.10 & 585 & 67.99 & 1127 & 52.03 & 1596 & 17.55 & 1433 & 58.03 & 1100 & \textbf{0.53} \\
\bottomrule
\end{tabular}%
}
\end{table*}

\subsection{Benchmarking Against Closed-Source State-of-the-Art Models}
\label{appendix:closed_source_comparison}

To situate the performance and efficiency of AdaThink-Med within the broader landscape of current large language models, we extended our evaluation to include GPT-5 and Gemini-2.5-Pro. It is important to acknowledge that these proprietary models are trained on significantly larger datasets and possess parameter counts orders of magnitude higher than the 7B/8B scale models used in our main experiments. For the purpose of the Accuracy-Efficiency Score (AES) calculation in this subsection, we utilized the performance of the Qwen2.5-Instruct-7B model (fine-tuned with GRPO) as the baseline reference.

The comparative results are detailed in Table~\ref{tab:closed_source_comparison}. We observe that while both closed-source models achieve high accuracy, their reasoning behaviors differ markedly. Gemini-2.5-Pro, despite its strong performance, tends to generate lengthy and often redundant reasoning chains, resulting in suboptimal inference efficiency. In contrast, GPT-5, which reportedly incorporates proprietary adaptive reasoning optimizations, demonstrates a significant reduction in output length while maintaining high accuracy.

Notably, AdaThink-Med (Llama backbone) achieves an inference efficiency profile comparable to that of GPT-5, reducing the average response length to just 64 tokens. While there remains a performance gap in absolute accuracy attributable to the vast disparity in model size (8B vs. hundreds of billions of parameters), the high AES scores obtained by our approach highlight its effectiveness. This demonstrates that AdaThink-Med successfully balances performance and computational cost, achieving state-of-the-art efficiency levels on smaller, open-weights models that are accessible for widespread deployment.

\begin{table*}[h]
\centering
\caption{Performance and efficiency comparison with closed-source state-of-the-art models (GPT-5 and Gemini-2.5-Pro). The Qwen-GRPO model serves as the baseline for AES calculations. AdaThink-Med achieves competitive efficiency scores despite the significant difference in model scale. Best AES scores are highlighted in \textbf{bold}.}
\label{tab:closed_source_comparison}
\resizebox{\textwidth}{!}{%
\begin{tabular}{lccccccccccccccc}
\toprule
\multirow{2}{*}{Model} & \multicolumn{2}{c}{MedQA} & \multicolumn{2}{c}{MedMCQA} & \multicolumn{2}{c}{PubMedQA} & \multicolumn{2}{c}{MMLU-ProM} & \multicolumn{2}{c}{GPQA-M} & \multicolumn{2}{c}{MedXpert} & \multicolumn{2}{c}{Avg} & \multirow{2}{*}{AES} \\
\cmidrule(lr){2-3} \cmidrule(lr){4-5} \cmidrule(lr){6-7} \cmidrule(lr){8-9} \cmidrule(lr){10-11} \cmidrule(lr){12-13} \cmidrule(lr){14-15}
 & Acc. & Len. & Acc. & Len. & Acc. & Len. & Acc. & Len. & Acc. & Len. & Acc. & Len. & Acc. & Len. & \\
\midrule
Qwen-GRPO & 67.55 & 504 & 59.52 & 378 & 72.30 & 396 & 66.12 & 535 & 48.46 & 575 & 14.48 & 592 & 54.74 & 497 & - \\
GPT-5 & 76.07 & 239 & 62.89 & 198 & 68.00 & 278 & 72.42 & 316 & 59.13 & 320 & 35.25 & 335 & 62.46 & 281 & \textbf{+0.86} \\
Gemini-2.5-Pro & 74.20 & 505 & 62.60 & 408 & 70.80 & 220 & 68.12 & 549 & 57.32 & 549 & 27.91 & 612 & 60.16 & 474 & +0.34 \\
AdaThink-Med [Llama] & 67.00 & 36 & 60.41 & 73 & 78.40 & 54 & 63.58 & 85 & 47.94 & 100 & 16.20 & 35 & 55.59 & 64 & \textbf{+0.92} \\
AdaThink-Med [Qwen] & 68.34 & 79 & 58.74 & 109 & 73.50 & 62 & 66.45 & 121 & 48.72 & 175 & 14.16 & 88 & 54.99 & 106 & +0.80 \\
\bottomrule
\end{tabular}%
}
\end{table*}

\subsection{Ablation Study on Batch Normalization and Batch Size}
\label{appendix:ablation_bn_batch}

To validate the architectural choices in AdaThink-Med, specifically the use of Batch Normalization (BN) and the impact of batch size, we conducted a series of ablation experiments. In our framework, BN serves two critical objectives: first, it constrains the output uncertainty to a normalized $[0, 1]$ interval, ensuring numerical stability; second, it works in conjunction with the Exponential Moving Average (EMA) to robustly estimate the global difficulty threshold of the dataset. This global threshold is pivotal as it dictates the aggressiveness of the adaptive compression—datasets identified as globally "simpler" allow for greater length reduction.

Table~\ref{tab:bn_batch_ablation} presents the comparative results of modifying batch normalization and batch size settings. Our analysis yields three key observations:

\begin{enumerate}
    \item \textbf{Necessity of Batch Normalization:} The removal of BN (denoted as \textit{AdaThink-Med-ablations(BN)}) leads to training instability. Without the normalization to a fixed range and the relative comparison provided by batch statistics, the algorithm fails to establish an accurate difficulty threshold. Consequently, the adaptive length penalty mechanism becomes ineffective, and the model's performance and output length revert to levels similar to the initial unoptimized model (AES +0.17), failing to achieve meaningful compression.
    
    \item \textbf{Impact of Batch Size on Convergence:} Comparing the standard configuration (Batch Size $256 \times 8$) with a reduced setting ($32 \times 8$), we observe that batch size primarily influences the convergence velocity rather than the final performance. Larger batch sizes provide more representative statistics for the difficulty estimator, accelerating convergence (200 steps). However, smaller batch sizes, while slower to converge (280 steps), eventually achieve comparable efficiency and accuracy (AES +0.88 vs +0.92).
    
    \item \textbf{Inefficacy of Per-Sample Calibration:} We also explored per-sample entropy calibration, which is theoretically equivalent to a batch size of 1. This approach resulted in high variance in difficulty estimation, leading to severe training instability and failure to converge. This confirms that the synergy between batch-level normalization and EMA is essential for robust global difficulty estimation and stable optimization.
\end{enumerate}

\begin{table*}[h]
\centering
\caption{Ablation study on Batch Normalization (BN) and Batch Size. "Conv. Steps" denotes the number of steps required for convergence. Removing BN leads to optimization failure, while smaller batch sizes result in slower convergence but comparable final efficiency.}
\label{tab:bn_batch_ablation}
\resizebox{\textwidth}{!}{%
\begin{tabular}{lcccccccccccccccccc}
\toprule
\multirow{2}{*}{Model} & \multirow{2}{*}{BN} & \multirow{2}{*}{Batch Size} & \multicolumn{2}{c}{MedQA} & \multicolumn{2}{c}{MedMCQA} & \multicolumn{2}{c}{PubMedQA} & \multicolumn{2}{c}{MMLU-ProM} & \multicolumn{2}{c}{GPQA-M} & \multicolumn{2}{c}{MedXpert} & \multicolumn{2}{c}{Avg} & \multirow{2}{*}{\shortstack{Conv.\\Steps}} & \multirow{2}{*}{AES} \\
\cmidrule(lr){4-5} \cmidrule(lr){6-7} \cmidrule(lr){8-9} \cmidrule(lr){10-11} \cmidrule(lr){12-13} \cmidrule(lr){14-15} \cmidrule(lr){16-17}
 & & & Acc. & Len. & Acc. & Len. & Acc. & Len. & Acc. & Len. & Acc. & Len. & Acc. & Len. & Acc. & Len. & & \\
\midrule
Llama-3.1-8B-Instruct [CoT] & - & - & 68.18 & 461 & 57.32 & 309 & 78.20 & 296 & 60.06 & 457 & 45.64 & 485 & 16.08 & 545 & 54.25 & 425 & - & - \\
Llama-3.1-8B-Instruct & - & - & 54.75 & 302 & 54.69 & 169 & 77.00 & 113 & 57.00 & 256 & 35.64 & 345 & 14.48 & 342 & 48.93 & 254 & - & -0.09 \\
Llama-3.1-8B-Instruct [GRPO] & - & - & 72.50 & 438 & 61.60 & 299 & 78.50 & 290 & 63.51 & 442 & 46.92 & 461 & 17.26 & 529 & 56.72 & 410 & - & +0.17 \\
AdaThink-Med-ablations (BN) & $\times$ & $256 \times 8$ & 70.43 & 425 & 59.33 & 267 & 78.20 & 275 & 62.55 & 429 & 46.49 & 423 & 17.00 & 500 & 55.76 & 387 & Fluct. & +0.17 \\
AdaThink-Med-ablations (BS) & $\checkmark$ & $32 \times 8$ & 66.98 & 40 & 60.21 & 82 & 78.90 & 60 & 62.65 & 93 & 46.53 & 96 & 15.76 & 42 & 55.04 & 69 & 280 & +0.88 \\
AdaThink-Med & $\checkmark$ & $256 \times 8$ & 67.00 & 36 & 60.41 & 73 & 78.40 & 54 & 63.58 & 85 & 47.94 & 100 & 16.20 & 35 & 55.59 & 64 & 200 & +0.92 \\
\bottomrule
\end{tabular}%
}
\end{table*}

\subsection{Validation of Dataset Selection Strategy against Random Sampling}
\label{appendix:dataset_selection_validation}

To verify that the performance gains observed in our dataset selection experiments stem from the intrinsic quality of the selected instances rather than merely the reduction in dataset size, we conducted a controlled comparison against random sampling. For each dataset retention ratio $\delta \in \{20\%, 40\%, 60\%, 80\%\}$, we constructed a control subset by randomly selecting samples from the full training corpus.

To ensure a fair comparison, we standardized the computational budget across all experiments by fixing the number of training steps at 200. This approach allows the model to achieve convergence regardless of the subset size, isolating the impact of data quality.

The results, presented in Table~\ref{tab:dataset_selection_control}, demonstrate that our proposed dataset selection method consistently outperforms random selection across all retention ratios. For example, at $\delta=40\%$, our method achieves an average accuracy of 55.12\%, surpassing the random baseline (54.20\%) and even slightly exceeding the full dataset baseline (54.74\%). These findings confirm that AdaThink-Med effectively identifies high-value training samples that contribute most significantly to reasoning capability, validating the efficacy of our selection strategy beyond simple data reduction.

\begin{table*}[h]
\centering
\caption{Comparison of AdaThink-Med's dataset selection strategy against random subset selection across various retention ratios ($\delta$). To ensure fairness, all experiments were conducted with a fixed compute budget of 200 training steps. Our method consistently outperforms random selection, indicating the superior quality of the selected data.}
\label{tab:dataset_selection_control}
\resizebox{\textwidth}{!}{%
\begin{tabular}{lccccccccccccccc}
\toprule
\multirow{2}{*}{\textbf{$\delta$}} & \multirow{2}{*}{\textbf{Method}} & \multicolumn{2}{c}{MedQA} & \multicolumn{2}{c}{MedMCQA} & \multicolumn{2}{c}{PubMedQA} & \multicolumn{2}{c}{MMLU-ProM} & \multicolumn{2}{c}{GPQA-M} & \multicolumn{2}{c}{MedXpert} & \multicolumn{2}{c}{Avg.} \\
\cmidrule(lr){3-4} \cmidrule(lr){5-6} \cmidrule(lr){7-8} \cmidrule(lr){9-10} \cmidrule(lr){11-12} \cmidrule(lr){13-14} \cmidrule(lr){15-16}
 & & Acc. & Len. & Acc. & Len. & Acc. & Len. & Acc. & Len. & Acc. & Len. & Acc. & Len. & Acc. & Len. \\
\midrule
0\% & - & 63.86 & 490 & 56.68 & 359 & 73.10 & 386 & 62.28 & 523 & 46.41 & 562 & 12.28 & 585 & 52.44 & 484 \\
\midrule
\multirow{2}{*}{20\%} & Random & 64.21 & 505 & 57.23 & 384 & 73.50 & 391 & 62.88 & 537 & 46.99 & 588 & 13.22 & 577 & 53.01 & 497 \\
 & Ours & 64.81 & 497 & 58.50 & 373 & 74.40 & 400 & 64.82 & 526 & 48.21 & 565 & 14.00 & 585 & 54.12 & 491 \\
\midrule
\multirow{2}{*}{40\%} & Random & 64.38 & 523 & 57.66 & 358 & 72.30 & 415 & 66.65 & 515 & 50.22 & 557 & 13.99 & 585 & 54.20 & 493 \\
 & Ours & 65.44 & 512 & 58.81 & 367 & 74.30 & 396 & 66.51 & 522 & 51.28 & 564 & 14.37 & 590 & 55.12 & 492 \\
\midrule
\multirow{2}{*}{60\%} & Random & 64.96 & 507 & 57.43 & 352 & 72.60 & 416 & 65.38 & 532 & 48.32 & 532 & 14.02 & 551 & 53.79 & 482 \\
 & Ours & 68.50 & 488 & 58.69 & 356 & 73.10 & 381 & 66.12 & 513 & 47.44 & 535 & 13.96 & 569 & 54.64 & 474 \\
\midrule
\multirow{2}{*}{80\%} & Random & 66.38 & 499 & 56.73 & 366 & 71.20 & 398 & 66.33 & 515 & 48.96 & 528 & 14.67 & 560 & 54.05 & 478 \\
 & Ours & 68.42 & 504 & 58.38 & 370 & 72.30 & 388 & 67.49 & 527 & 50.00 & 576 & 14.73 & 579 & 55.22 & 491 \\
\midrule
100\% & - & 67.55 & 504 & 59.52 & 378 & 72.30 & 396 & 66.12 & 535 & 48.46 & 575 & 14.48 & 592 & 54.74 & 497 \\
\bottomrule
\end{tabular}%
}
\end{table*}

\begin{table*}[htbp]
\caption{Dataset selection experimental results. $\delta$ represents the ratio of training datasets.
}
\label{tab:delta_performance}
\centering
\resizebox{\textwidth}{!}{%
\begin{tabular}{lcc|cc|cc|cc|cc|cc|cc}
\toprule
\multirow{3}{*}{\textbf{$\delta$}} &
\multicolumn{4}{c|}{\textbf{In-domain}} &
\multicolumn{8}{c|}{\textbf{Out-domain}} &
\multicolumn{2}{c}{\textbf{Avg.}} \\
\cmidrule(lr){2-5} \cmidrule(lr){6-13} \cmidrule(lr){14-15}
 & \multicolumn{2}{c}{MedQA} & \multicolumn{2}{c|}{MedMCQA} 
 & \multicolumn{2}{c}{PubMedQA} & \multicolumn{2}{c}{MMLU-ProM} & \multicolumn{2}{c}{GPQA-M} & \multicolumn{2}{c|}{MedXpert} & \multirow{2}{*}{Acc.} & \multirow{2}{*}{Len.} \\
\cmidrule(lr){2-3} \cmidrule(lr){4-5} \cmidrule(lr){6-7} \cmidrule(lr){8-9} \cmidrule(lr){10-11} \cmidrule(lr){12-13}
 & Acc. & Len. & Acc. & Len. & Acc. & Len. & Acc. & Len. & Acc. & Len. & Acc. & Len. & & \\
\midrule
0\%   & 63.86 & 490 & 56.68 & 359 & 73.10 & 386 & 62.28 & 523 & 46.41 & 562 & 12.28 & 585 & 52.44 & 484 \\
20\%  & 64.81 & 497 & 58.50 & 373 & 74.40 & 400 & 64.82 & 526 & 48.21 & 565 & 14.00 & 585 & 54.12 & 491 \\
40\%  & 65.44 & 512 & 58.81 & 367 & 74.30 & 396 & 66.51 & 522 & 51.28 & 564 & 14.37 & 590 & 55.12 & 492 \\
60\%  & 68.50 & 488 & 58.69 & 356 & 73.10 & 381 & 66.12 & 513 & 47.44 & 535 & 13.96 & 569 & 54.63 & 474 \\
80\%  & 68.42 & 504 & 58.38 & 370 & 72.30 & 388 & 67.49 & 527 & 50.00 & 576 & 14.73 & 579 & 55.22 & 491 \\
100\% & 67.55 & 504 & 59.52 & 378 & 72.30 & 396 & 66.12 & 535 & 48.46 & 575 & 14.48 & 592 & 54.74 & 497 \\
\bottomrule
\end{tabular}
}
\end{table*}

\section{Qualitative Results}
\subsection{Case Studies}
\begin{figure*}
    \centering
    \includegraphics[width=1\linewidth]{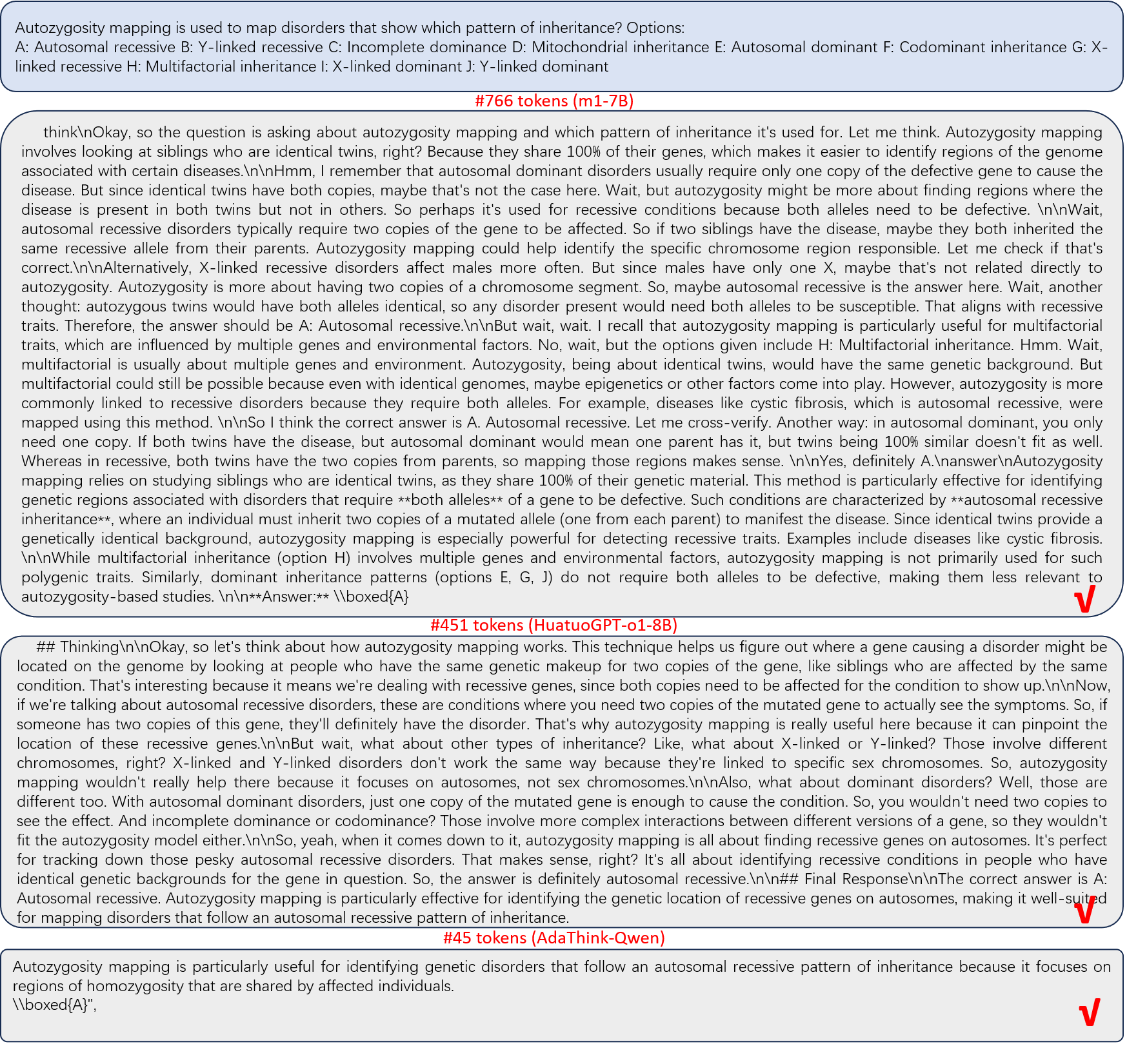}
    \caption{Case 1. AdaThink-Med delivers correct answer while consumes far fewer tokens than m1-7B and HuatuoGPT-o1-8B.}
    \label{fig:case_1}
\end{figure*}
Fig.~\ref{fig:case_1} presents a straightforward evaluation of basic genetics knowledge. The baseline models, m1 and HuatuoGPT-o1, display evident overthinking through numerous redundant reasoning steps, consuming 766 and 451 tokens respectively. In contrast, AdaThink-Med achieves a correct solution with only 45 tokens, demonstrating conciseness and accuracy. Furthermore, the baselines frequently produce incorrect conclusions during analysis and rely on a ``wait" mechanism to initiate unnecessary rethinking cycles.

\begin{figure*}
    \centering
    \includegraphics[width=1\linewidth]{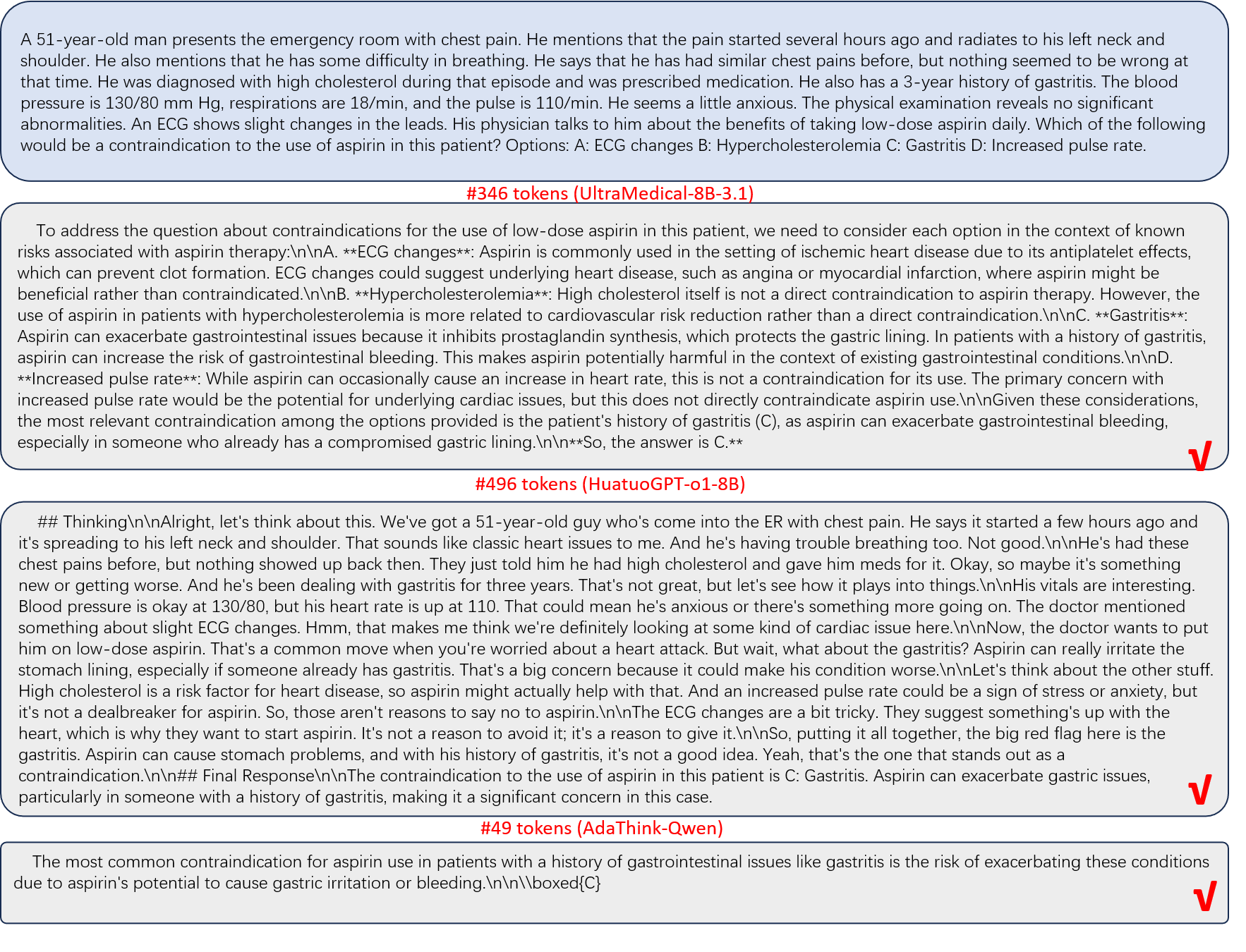}
    \caption{Case 2. AdaThink-Med delivers correct answer while consumes far fewer tokens than UltraMedical-8B-3.1 and HuatuoGPT-o1-8B.}
    \label{fig:case_2}
\end{figure*}

\begin{figure*}
    \centering
    \includegraphics[width=1\linewidth]{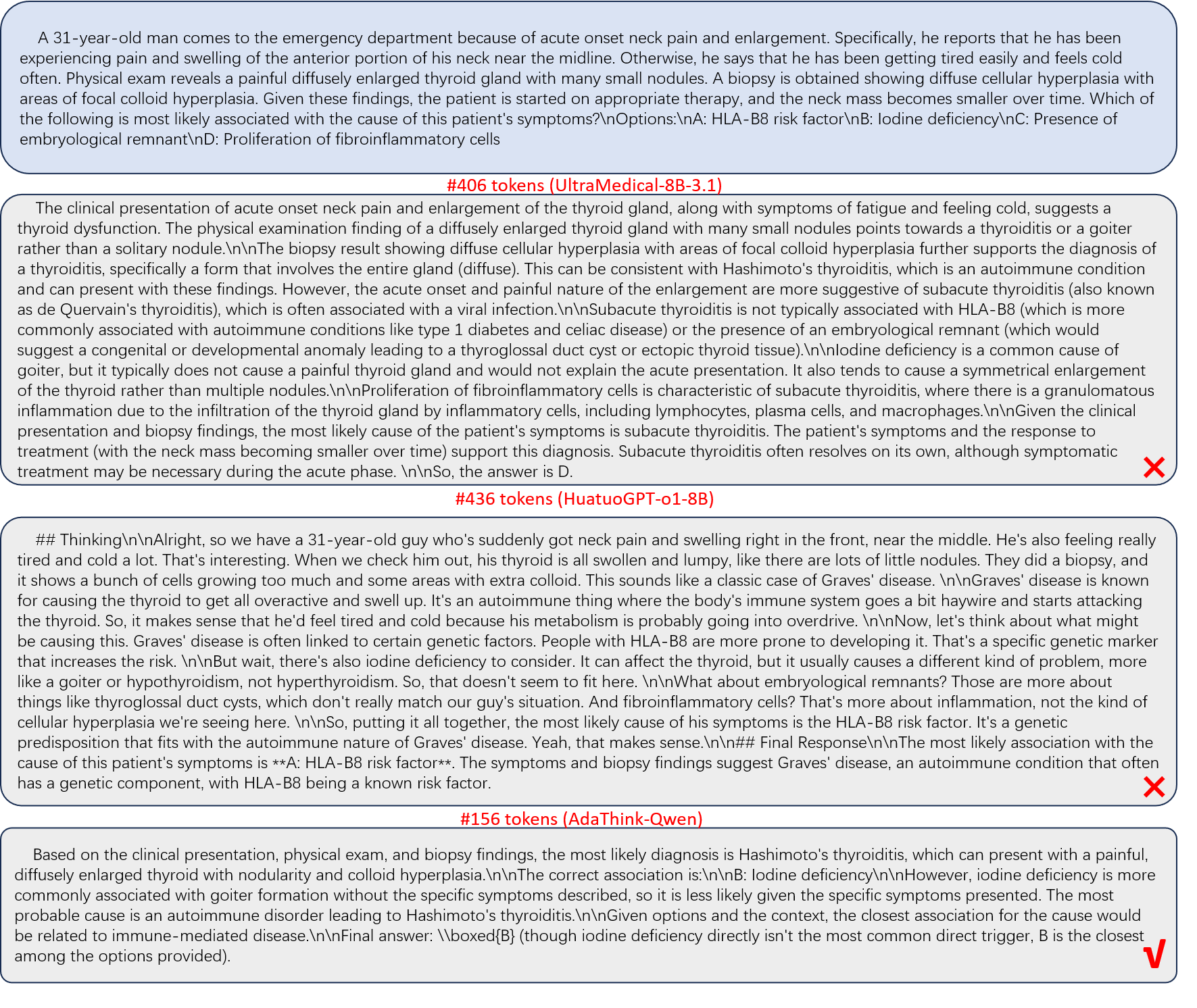}
    \caption{Case 3. An overthinking case of UltraMedical-8B-3.1 and HuatuoGPT-o1-8B.}
    \label{fig:case_3}
\end{figure*}

\begin{figure*}
    \centering
    \includegraphics[width=1\linewidth]{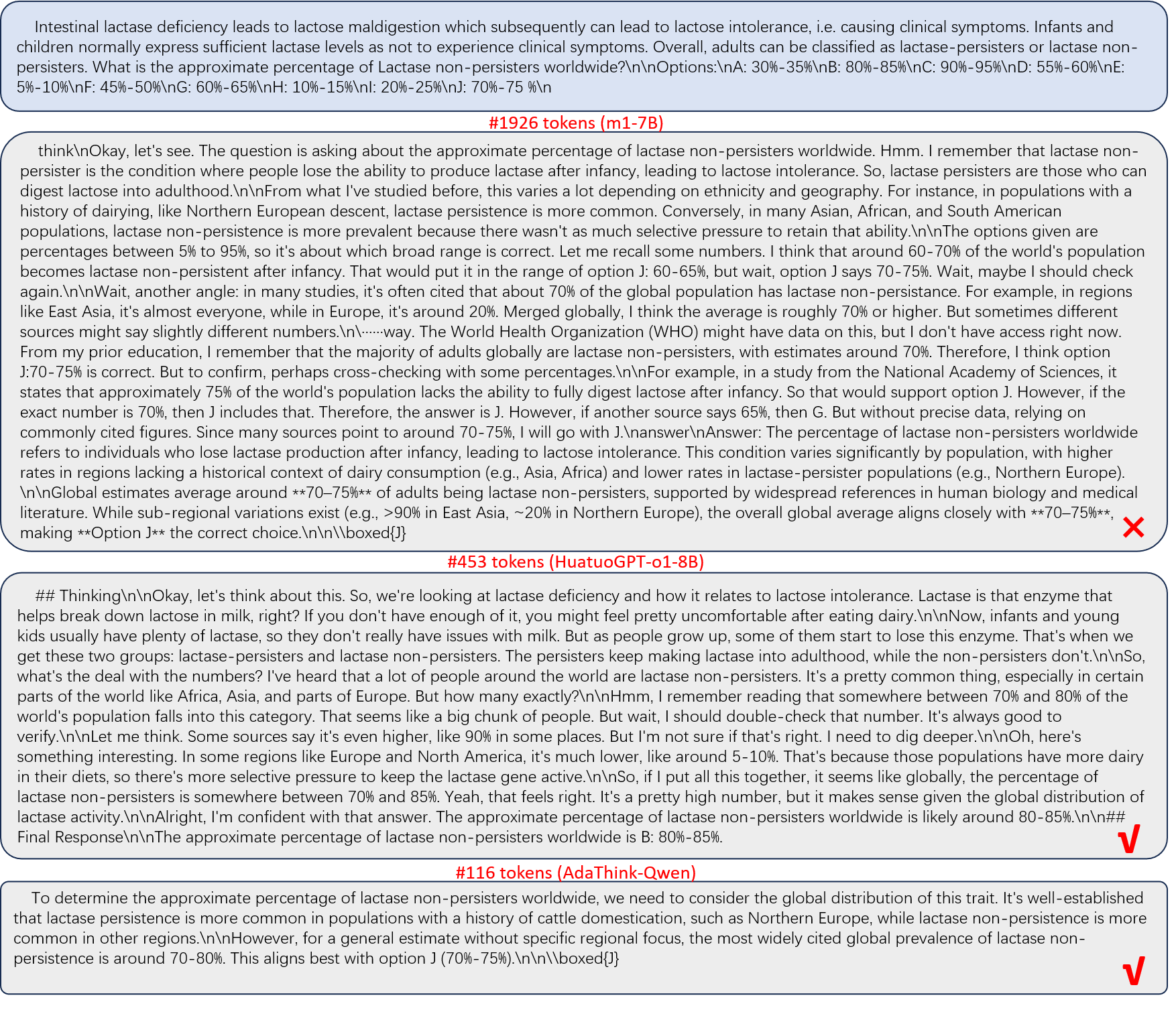}
    \caption{Case 4. An overthinking case of HuatuoGPT-o1-8B.}
    \label{fig:case_4}
\end{figure*}

\begin{figure*}[t]
    \centering
    \includegraphics[width=0.8\linewidth]{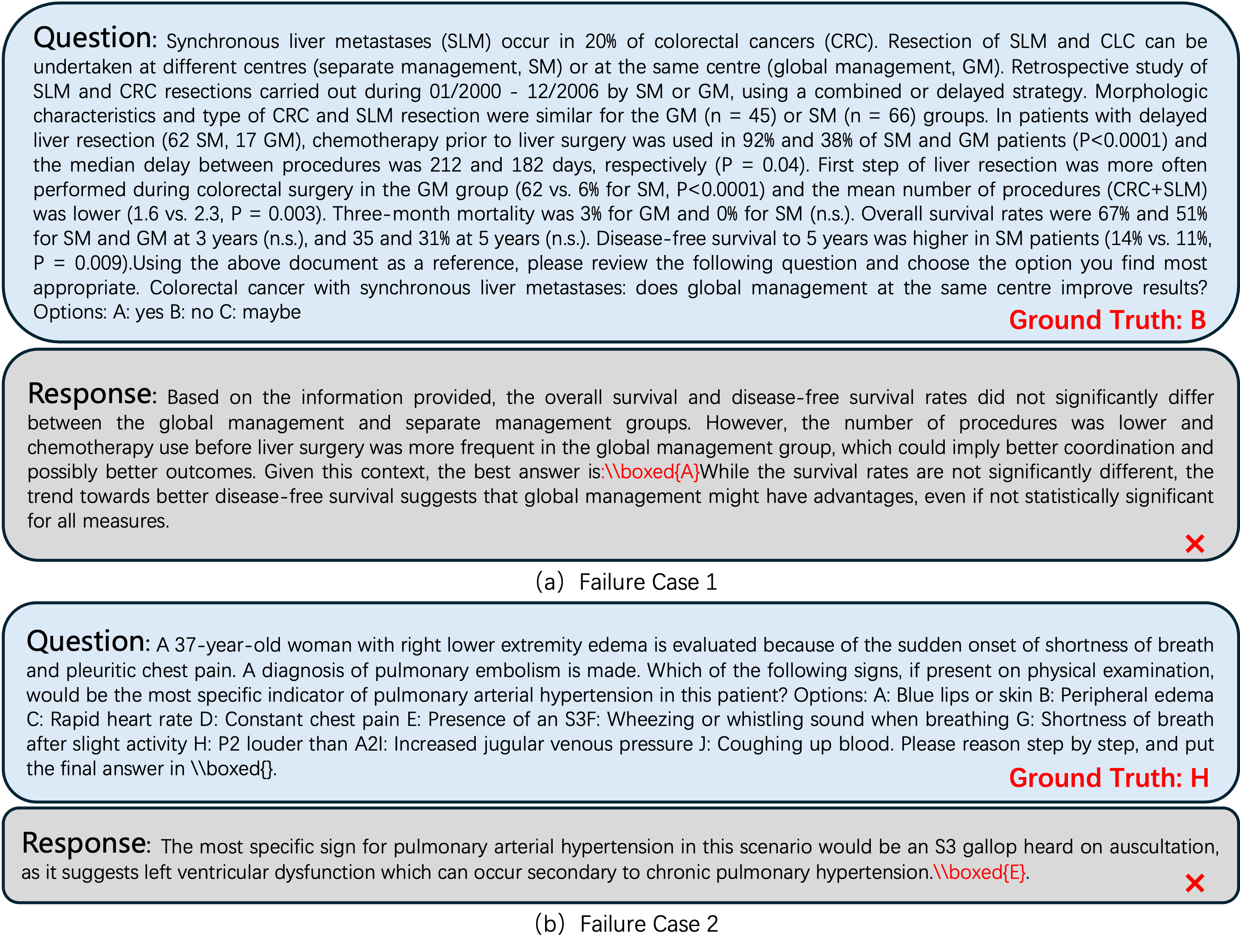}
    \caption{{Failure Cases Study}}
    \label{fig:fail_case}
\end{figure*}

\subsection{Failure Case Studies}
Upon careful examination, we identified that for medium-difficulty tasks, the model struggled with balancing non-thinking and thinking, often relying on quick thinking that resulted in insufficient reasoning. In failure case (a), the model's output incorrectly suggests that global management might improve outcomes based on trends toward better disease-free survival, despite the lack of statistically significant differences in overall survival and disease-free survival between global management and separate management. This reflects a failure in properly interpreting the study results, where no significant differences in survival were found, and the disease-free survival advantage for SM was statistically significant. The model appears to overemphasize minor trends, neglecting the statistical evidence that showed global management did not provide superior results in the key outcomes. This oversight underscores the importance of distinguishing between statistical significance and potential clinical trends, particularly when the data does not support a definitive advantage for one management strategy over the other. Consequently, a more nuanced approach should have been employed to align the analysis with the study's findings. In failure case (b), the model incorrectly identifies the S3 gallop (E) as the most specific sign for pulmonary arterial hypertension. While an S3 sound can indicate left ventricular dysfunction, it is more commonly associated with volume overload or heart failure, not specifically PAH. In contrast, the sign of P2 louder than A2 (H) is far more specific to pulmonary arterial hypertension, as it directly reflects increased pressure in the pulmonary arteries. The model failed to consider this crucial auscultatory finding, which is the hallmark sign of PAH, particularly as the increased pulmonary pressure causes delayed and louder closure of the pulmonary valve (P2). Thus, the model did not appropriately prioritize the most specific sign for PAH.
\end{document}